\documentclass[twoside,11pt]{article}
\usepackage{dmlr2e}

\usepackage{amsmath,amsfonts,bm}

\def\eqref#1{equation~\ref{#1}}

\def\1{\bm{1}}

\DeclareMathAlphabet{\mathsfit}{\encodingdefault}{\sfdefault}{m}{sl}
\SetMathAlphabet{\mathsfit}{bold}{\encodingdefault}{\sfdefault}{bx}{n}

\usepackage{hyperref}
\usepackage{url}

\usepackage{makecell}
\newcommand{\prcr}[2]{\makecell{#1\\{\scriptsize(#2)}}}

\usepackage[most]{tcolorbox}
\tcbuselibrary{listingsutf8}
\usepackage{minted}
\usepackage{caption}
\usepackage[utf8]{inputenc} 
\usepackage[T1]{fontenc}    
\usepackage{lmodern}         
\usepackage{hyperref}       
\usepackage{url}            
\usepackage{xurl}           
\usepackage{booktabs}       
\usepackage{amsfonts}       
\usepackage{nicefrac}       
\usepackage{microtype}      
\usepackage{xcolor}         
\usepackage[textsize=tiny]{todonotes}
\usepackage{enumitem}
\usepackage{multirow}       
\usepackage{graphicx}       
\usepackage{subfigure}

\usepackage{colortbl}       
\usepackage{pifont}
\usepackage{multicol}
\usepackage{array}
\usepackage{wrapfig}
\usepackage{tikz}
\usepackage{xspace}
\usepackage{cleveref}
\usetikzlibrary{shadows, positioning}

\usepackage[most]{tcolorbox}
\usepackage{enumitem}
\definecolor{takeawaybg}{RGB}{238,238,248}
\definecolor{takeawayfg}{RGB}{76,79,178}
\definecolor{summarybg}{RGB}{234,245,236}
\definecolor{summaryfg}{RGB}{40,120,60}

\newtcolorbox{takeaway}[1][Takeaway]{enhanced, frame hidden, boxrule=0pt, arc=8pt,
  colback=takeawaybg, left=10pt, right=10pt, top=8pt, bottom=8pt,
  before upper={\textcolor{takeawayfg}{\sffamily\bfseries #1:}\ }}
\newtcolorbox{summarybox}[1]{enhanced, frame hidden, boxrule=0pt, arc=8pt,
  colback=summarybg, left=10pt, right=10pt, top=8pt, bottom=8pt,
  before upper={\textcolor{summaryfg}{\sffamily\bfseries #1:}\ }}

\definecolor{darkgreen}{rgb}{0.0, 0.5, 0.0}
\definecolor{verylightgray}{rgb}{0.97, 0.97, 0.97}

\newcommand{\benchmark}{\textsc{LLM-BabyBench}\xspace}

\title{LLM-BabyBench: Can Language Models Plan in Worlds They Can Simulate?}

\author{\name Idriss Malek$^{*\dagger}$ \email im2868@nyu.edu\\
      \addr Tandon School of Engineering\\
      New York University
      \AND
      \name Omar Choukrani$^{*}$ \email omar.choukrani@mbzuai.ac.ae \\
      \addr Mohamed bin Zayed University of Artificial Intelligence (MBZUAI)\\
      Abu Dhabi, UAE
      \AND
      \name Daniil Orel$^{*}$ \email daniil.orel@mbzuai.ac.ae \\
      \addr Mohamed bin Zayed University of Artificial Intelligence (MBZUAI)\\
      Abu Dhabi, UAE
            \AND
      \name Anh Duy Le Dinh \email anhduy.ledinh@mbzuai.ac.ae \\
      \addr Mohamed bin Zayed University of Artificial Intelligence (MBZUAI)\\
      Abu Dhabi, UAE
      \AND
      \name Zhuohan Xie \email zhuohan.xie@mbzuai.ac.ae \\
      \addr Mohamed bin Zayed University of Artificial Intelligence (MBZUAI)\\
      Abu Dhabi, UAE
      \AND
      \name Zangir Iklassov \email zangir.iklassov@mbzuai.ac.ae \\
      \addr Mohamed bin Zayed University of Artificial Intelligence (MBZUAI)\\
      Abu Dhabi, UAE
      \AND
      \name Martin Tak\'a\v{c} \email martin.takac@mbzuai.ac.ae \\
      \addr Mohamed bin Zayed University of Artificial Intelligence (MBZUAI)\\
      Abu Dhabi, UAE
      \AND
      \name Salem Lahlou \email salem.lahlou@mbzuai.ac.ae \\
      \addr Mohamed bin Zayed University of Artificial Intelligence (MBZUAI)\\
      Abu Dhabi, UAE}

\def\openreview{\url{https://openreview.net/forum?id=XXXX}} 
\ShortHeadings{LLM-BabyBench: Can Language Models Plan in Worlds They Can Simulate?}{Malek et al.}
\firstpageno{1}
\editor{}

\begin{document}

\maketitle

\begingroup
\renewcommand{\thefootnote}{\fnsymbol{footnote}}
\footnotetext[1]{Equal contribution.}
\footnotetext[2]{Work partly done while at MBZUAI.}
\endgroup
\begin{abstract}
Interactive benchmarks for LLMs usually report a single metric, even though succeeding in them requires models to show several different skills. While those benchmarks are useful for measuring the overall capability of models to perform the downstream task, it is harder to identify which skill was lacking and caused failure solely from the aggregate score. For example, in benchmarks where models are supposed to navigate a simulated world to perform a task, failures can come from perception, an ambiguous instruction, retrieval, missing commonsense about what actions do, unknown dynamics, or planning. \benchmark recasts the procedurally generated BabyAI gridworld as a fully observable, purely textual environment designed to evaluate planning separately from those factors. The nature of the environment and the structure of the prompt control for the other factors cited above. To evaluate the models, we define the PPD suite: \textbf{Predict} asks for the state that follows an action sequence, \textbf{Plan} asks for an action sequence that reaches a goal, and \textbf{Decompose} asks for a subgoal sequence that achieves a mission. Across seven frontier and open models, queried zero-shot, we find simulation far ahead of planning for every model, close to saturated for the strongest. The length of the required solution governs planning failure, and each of the four strongest models has a horizon beyond which single-attempt success collapses. The responses of claude-opus-4-6 and claude-sonnet-4-6, which show their working step by step, reveal that they plan by proposing L-shaped corridors, which is consistent with their successful plans staying near-optimal while success falls. At the subgoal level, decomposition precision likewise falls to zero on long missions, while comprehension of those missions persists.

\end{abstract}


\section{Introduction}
Language models are increasingly used as agents that act over many steps, for example to call tools, browse the web, control robots or run multi-stage workflows.
To evaluate the performance of LLMs in simulated low-stakes environments, many interactive benchmarks have been developed. Usually, those benchmarks assess the models' performance through a single metric that represents the overall success on the task~\citep{ALFWorld20, savva2019habitat, fan2022minedojo, liu2024agentbench, paglieri2025balrog}. While this aggregated measure is a fair summary of overall performance in a particular setting, it is a poor basis for a claim about any single capability, especially planning~\citep{li2024embodiedagentinterface}. For example, an agent failing to fetch a key in a household simulator can be due to perception, an ambiguous instruction, retrieval from a long context, lack of commonsense knowledge, or a misunderstanding of the environment's transition dynamics. Studies of planning conclude that LLMs cannot plan autonomously~\citep{valmeekam2023planning, kambhampati2024llmmodulo}, and the collapse of reasoning models on long puzzles has been read both as a limit of reasoning~\citep{shojaee2025illusion} and as errors in executing a plan the model already knows~\citep{sinha2026longhorizon}. Such conflicting readings may be due to the tendency to treat overall performance as an indicator of planning capability.

The goal of \benchmark is to evaluate planning while holding the other factors fixed. We transform the BabyAI gridworld \citep{babyai, MinigridMiniworld23} into a purely textual and fully observable environment. The instructions are written in Baby Language, a small formal grammar, which guarantees that the missions themselves are unambiguous; one ambiguity in how our prompt defines a subgoal argument is documented in Appendix~\ref{apx:convention}. The prompt contains all the information needed to solve the problems, which avoids failures due to retrieval, perception, or misunderstanding of the world's rules. The environment also allows programmatic validation, which avoids relying on LLM-as-a-judge and the bias that comes with it~\citep{ye2025justice}.

We also introduce a novel evaluation protocol, named PPD. \textbf{Predict} asks the model to predict the final state given an initial state and an action sequence. \textbf{Plan} expects an action sequence that leads to a specific final state, and measures both success rate and the efficiency of the trajectory provided. \textbf{Decompose} supplies a high-level mission and asks for a sequence of subgoals that achieves it when executed.

Our contributions are:
\begin{itemize}
\item \textbf{\textsc{LLM-BabyBench}}, a fully observable textual gridworld, and the \textbf{PPD suite}, a standardised evaluation harness to assess planning capabilities of LLMs.
\item \textbf{The OmniBot}, a deterministic, full-visibility expert that solves every instance and is used as a verifier to evaluate LLMs.
\item \textbf{New levels} with variable densities and sizes, specifically tailored for the \textbf{Plan} task. They hold task type, action set and goal structure fixed, which lets us show that completion tracks the length of the solution rather than the size of the grid.
\item \textbf{Three assistance-aware metrics} to evaluate the \textbf{Decompose} task, which measure both whether a model's decomposition succeeds and how much it helps a verified solver.
\item \textbf{An evaluation of seven frontier and open models} showing that simulation accuracy is close to saturated for the strongest models while planning completion collapses at long horizons.
\end{itemize}
The benchmark instances are available at \url{https://huggingface.co/datasets/salem-mbzuai/LLM-BabyBench} under the MIT license (24{,}000 instances: 3 tasks $\times$ 16 levels $\times$ 500 seeds; the experiments reported here use a 20-seed subset of each level; the supplementary datasheet documents composition, hosting and maintenance), and the environment, the OmniBot and the evaluation harness at \url{https://github.com/Idriss-Malek/LLM-BabyBench}.

\section{Related Work}
\paragraph{Interactive benchmarks for LLM agents.} Many benchmarks have been developed to assess different capabilities of LLMs. In particular, there is a large corpus of interactive benchmarks that evaluate navigation and task completion in simulated worlds. ALFRED \citep{alfred} and ALFWorld \citep{ALFWorld20} pair household instructions with embodied or text-based execution. Habitat \citep{savva2019habitat} evaluates photorealistic navigation. TextWorld \citep{cote2018textworld} and ScienceWorld \citep{wang2022scienceworld} evaluate LLMs on text games. MineDojo \citep{fan2022minedojo} and Voyager \citep{wang2023voyager} use Minecraft to test LLMs' behaviour in open worlds. There are also several suites that aggregate many tests under one harness: AgentBench \citep{liu2024agentbench}, SmartPlay \citep{wu2024smartplay}, BALROG \citep{paglieri2025balrog}, Crafter \citep{hafner2022crafter}, MiniHack \citep{samvelyan2021minihack}, and NetHack \citep{kuttler2020nethack}. Closest to our setting, BabyAI-Text \citep{grounding_online_rl} renders BabyAI observations as text, and BALROG \citep{paglieri2025balrog} uses it to evaluate LLM agents: the model reads a partial, egocentric description, acts one step at a time, and each episode is scored on completion. \benchmark instead gives the full grid with absolute coordinates, asks for a complete plan in a single response, and splits the evaluation into the three PPD tasks.

\paragraph{Planning benchmarks.} A second family of benchmarks was designed to evaluate planning directly. PlanBench \citep{planbench} and the accompanying analysis \citep{valmeekam2023planning} generate problems in domains from the International Planning Competition (IPC) and validate them using formal rules. Even for reasoning models, the autonomous generation rate collapses on long-horizon tasks~\citep{valmeekam2025o1}. \textsc{Natural Plan} \citep{naturalplan} and TravelPlanner \citep{xie2024travelplanner} evaluate LLMs in scheduling and itinerary domains, where extracting constraints from long tool outputs is necessary for planning. PPNL~\citep{aghzal2023ppnl} evaluates grid path planning in language and studies grid size, obstacle count and distance to the goal, observing that difficulty is temporal rather than spatial for its strongest configuration. Its follow-up~\citep{aghzal2024lookfurther} samples instances by optimal path length on maze-like $25\times25$ grids, finds that success falls with path length and depends on the geometry, and reports that $15\times15$ mazes are easier than $25\times25$ ones at the same path lengths, the smaller mazes also having fewer obstacles. We hold obstacle count fixed instead: stratifying by solution length within a fixed grid size reproduces the collapse, and at matched obstacle count the larger, less dense grid is easier for the three strongest models (Appendix~\ref{apx:plan-analysis}). Finally, CogEval \citep{momennejad2023cogeval} evaluates planning over graph-structured cognitive maps and documents hallucinations and loops.

\paragraph{Factoring evaluation into abilities.} Embodied Agent Interface~\citep{li2024embodiedagentinterface} decomposes embodied decision making into goal interpretation, subgoal decomposition, action sequencing and transition modelling, using VirtualHome and BEHAVIOR as testbeds. Several agent architectures~\citep{song2023llmplanner, ahn2022saycan} rely implicitly on the same separation between high-level planners and low-level actors. The PPD suite differs in that all three abilities are measured in one procedurally generated world against one deterministic oracle, so a model's prediction accuracy and its planning performance refer to the same dynamics and the same ground truth. Embodied Agent Interface also refines LLM subgoals into actions with a search-based planner, but scores whether the resulting trajectory is feasible and reaches the goal; we instead let a deterministic solver insert the subgoals the model omitted and measure success by how many it has to add.

\paragraph{Probing world models.} \citet{vafa2024worldmodel} show that generative models can hold incorrect world models even if they perform well on next-token prediction. \citet{li2023othello} and \citet{gurnee2024space} found that those models can still hold coherent internal representations of board state and of space and time. \citet{hao2023rap} even manage to use an LLM's own world model as a simulator inside search. We make no claim about internal representations. The \textbf{Predict} task only checks empirically that predicting and simulating the dynamics of the world are not the limiting factors for planning.

\paragraph{LLMs as components of planning systems.}
There is a substantial literature about LLMs being used for a fast approximate solution before an exact oracle verifies and completes it: affordance-grounded skill selection \citep{ahn2022saycan}, translation of free-form plans into admissible actions \citep{huang2022zeroshotplanners}, closed-loop replanning from environment feedback \citep{huang2022innermonologue}, program- and code-structured plans \citep{liang2023codeaspolicies, singh2023progprompt}, translation into a formal problem for a classical planner \citep{liu2023llmp}, and few-shot grounded planning for embodied agents \citep{song2023llmplanner}. \citet{kambhampati2024llmmodulo} argues that LLMs cannot plan autonomously and should instead generate candidates for sound external verifiers. A parallel line of work trains sequence models on search traces so that the model itself learns to search~\citep{lehnert2024searchformer, gandhi2024streamofsearch}. In the \textbf{Decompose} task, the Assistance Curve Integral tests whether the solutions provided by LLMs are worth refining. The subgoals themselves follow the standard notion of a temporally extended action \citep{sutton1999options}.

\section{\textsc{LLM-BabyBench} and the PPD suite}
\subsection{\textsc{LLM-BabyBench} and OmniBot}
\label{sec:omnibot}
We build on BabyAI \citep{babyai}, a procedurally generated grid world over the MiniGrid platform \citep{MinigridMiniworld23}. In this environment, an agent has to perform predefined tasks that are written in a small formal language named Baby Language. Baby Language combines verbs (go to, pick up, open, put next to), objects (door, key, ball, box), colours and relative locations (on your left, behind you) to provide simple instructions. The agent can open doors and pick up, carry and drop keys, balls and boxes. All items are colour-coded. MiniGrid defines seven actions (left, right, forward, pickup, drop, toggle, done); our textual interface exposes six, dropping \texttt{done}. BabyAI was built to study how Reinforcement Learning and Imitation Learning algorithms learn to follow grounded language instructions. To assess LLMs and their planning abilities, we adapt BabyAI to a purely textual, fully observable interface. Unlike the original BabyAI agent, the model receives all object positions regardless of its location or orientation. Except for the \textbf{Plan} task, for which we designed specific levels, the two other tasks use the original BabyAI levels.

Giving full visibility to the agent alters what the optimal path is, as there is no longer any need for blind exploration in this new setting. In particular, the Explore Subgoal is no longer needed. Full visibility also changes how locked doors are handled: the original bot can only fetch a matching key it has already seen and must otherwise explore for it, whereas the OmniBot sees every key and goes straight to it. Together with the exclusion of three levels that can generate unsolvable instances (Appendix~\ref{apx:levels}), this is what allows it to solve every instance we include. To produce ground truth, we extend BabyAI's hand-coded expert into an ``OmniBot'' that exploits full visibility to choose near-optimal paths. The OmniBot solves $100\%$ of benchmark instances and plays two roles: it supplies near-optimal trajectories (for the efficiency metrics) and it can execute an LLM's subgoals, optionally inserting its own (for the \textbf{Decompose} metrics, see Section~\ref{subsec:ppd}). While BabyAI's original bot was designed to be initialised with the original instruction, OmniBot can be initialised with any sequence of subgoals. The bot then appends whatever else the episode turns out to require. The number of subgoals added by the OmniBot is in particular used to assess the LLMs' own decomposition. 
The size of the oracle's own from-scratch decomposition of the mission, that is its initial plan plus every subgoal it inserts at runtime (the oracle subgoal count of Table~\ref{tab:decompose-by-count}), is the quantity $M_{\text{help}}$ that normalises the ACI metric below.

\textsc{LLM-BabyBench} is a minimal environment with clear and simple dynamics. That simplicity allows us to separate spatial planning from:
\begin{itemize}
\item \textbf{Perception}: partial observability and visual parsing (ALFRED~\citep{alfred}, Habitat~\citep{savva2019habitat}) can make a perception failure look like a planning failure. In \benchmark, the complete grid is given as text.
\item \textbf{Instruction ambiguity}: free-form goals (\textsc{Natural Plan}~\citep{naturalplan}, ALFWorld~\citep{ALFWorld20}) conflate planning with interpretation. Baby Language is a small formal grammar with templated instructions.
\item \textbf{Retrieval}: where planning depends on facts buried in long context or tool output (\textsc{Natural Plan}~\citep{naturalplan}), a failure may be a retrieval failure. In \benchmark, every object's coordinate is stated explicitly.
\item \textbf{World knowledge}: open-world tasks (MineDojo~\citep{fan2022minedojo}) require commonsense about what actions afford to perform planning. In \benchmark, all six actions and their deterministic effects are given in the prompt.
\item \textbf{Dynamics learning}: when transitions are unknown, failure may lie in learning the rules rather than in using them. In \benchmark, all dynamics are stated explicitly.
\item \textbf{Verification}: LLM-judged or self-reported success is can be gamed and is noisy. In \benchmark, verification is done by the OmniBot, which is deterministic and independent.
\end{itemize}
\subsection{The PPD suite}
\label{subsec:ppd}
\begin{figure}[htbp]
    \centering
    \includegraphics[width=.8\textwidth]{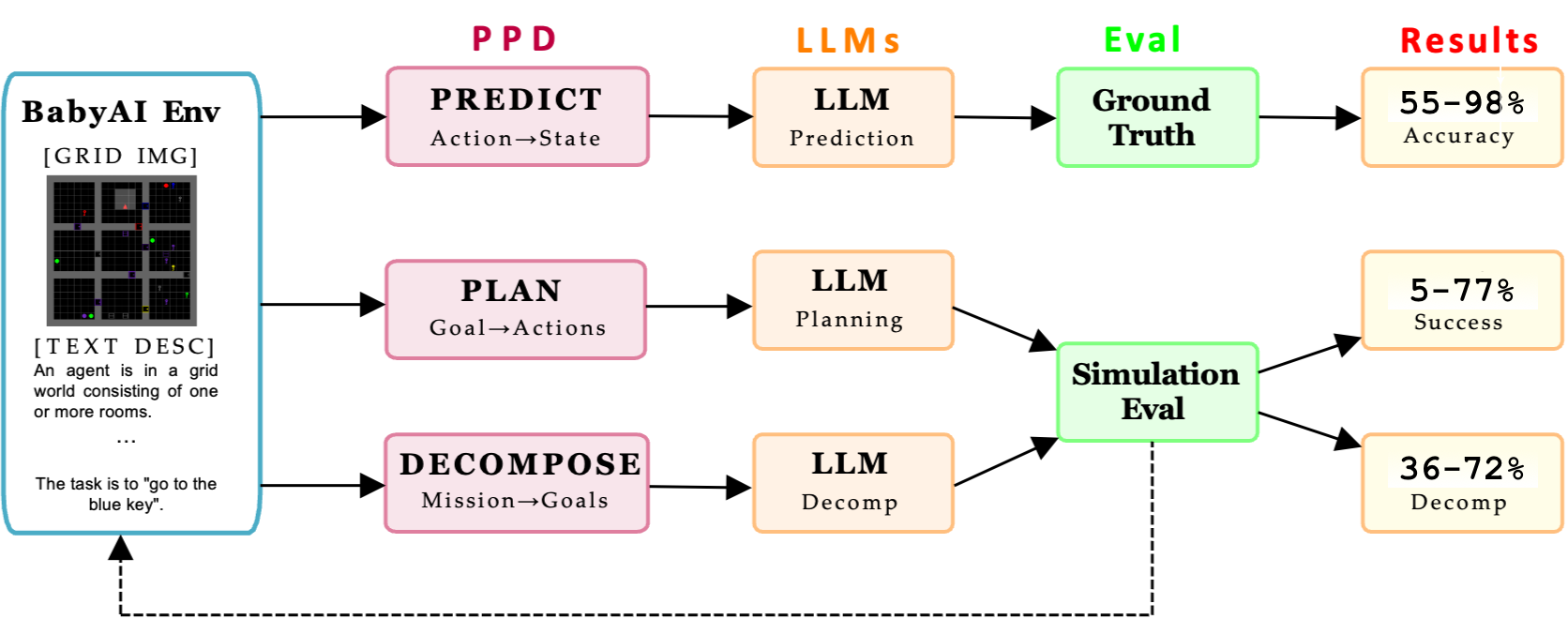}
    \caption{\textbf{The \benchmark PPD framework.} A BabyAI instance is serialised into a textual description that states the full grid, every object's coordinates, and the mission. We evaluate models using three tasks: \textbf{Predict} (action sequence $\rightarrow$ final state), \textbf{Plan} (goal $\rightarrow$ action sequence) and \textbf{Decompose} (mission $\rightarrow$ subgoal sequence).}
    \label{fig:main_framework}
\end{figure}

In order to use \textsc{LLM-BabyBench} to assess LLMs, we introduce the PPD (\textbf{Predict}, \textbf{Plan}, \textbf{Decompose}) suite:
\begin{itemize}
    \item \textbf{Predict}: given the initial state and an action sequence, output the final state (position and orientation). This task evaluates whether the model can track the \textsc{LLM-BabyBench} transition rules over an action sequence.
    \item \textbf{Plan}: given the initial state and a navigation goal (GoNextTo a target object), output an action sequence that reaches it. Correctness is evaluated by executing the sequence, and the LLM output is compared to the OmniBot's near-optimal solution to assess efficiency\footnote{The OmniBot is a strong but not always strictly optimal reference: on rare instances a model finds a shorter valid path, so measured efficiency can marginally exceed 1.}.
    \item \textbf{Decompose}:
    given the initial state and a high-level mission, output a subgoal sequence (from GoNextTo, Open, Close, Pickup, Drop) that, when executed, achieves the mission.
\end{itemize}
Figure~\ref{fig:main_framework} provides a conceptual overview of our PPD framework.
On \textbf{Predict} we use success rate. On \textbf{Plan} we use success rate together with efficiency, which compares the length of the OmniBot's own sequence with the number of actions the model used.\\
\textbf{Decompose} requires more care, since subgoals, unlike actions, are not directly executable. Simply comparing the sequence output by LLMs with the subgoal sequence computed by the OmniBot would be misleading, as a valid subgoal sequence is not unique. We therefore rely on the OmniBot to translate the LLM's subgoals into actions, adding any necessary intermediate subgoals along the way. All the metrics we rely on to assess LLMs on \textbf{Decompose} are based on how many additional subgoals the OmniBot needs to add to the output of the LLMs to complete the mission. The OmniBot, however, is never given the final goal. If the LLM fails to break the original task down into at least its essential subgoals, not even the OmniBot can supply the missing steps. Given the task \emph{Open a green door then pick the red ball}, for instance, the LLM must produce at least \emph{Open green door} and \emph{Pick red ball}.
The Comprehension Rate (CR) measures the success rate when the OmniBot is allowed to add as many subgoals as needed; it captures whether the LLM understood the subgoal logic and the final goal. This metric is essential to isolate planning. For models that have low CR, it is difficult to judge whether the failure is in planning or instruction understanding. The Precision Rate (PR) measures the success rate with no help from the OmniBot. A large gap between CR and PR shows that the model understands what to do but is unable to write a fully correct decomposition. Finally, the Assistance Curve Integral (ACI) measures how much help the model needs from the OmniBot. We integrate success over the number of subgoals the bot is permitted to insert, from $0$ to the oracle's own from-scratch decomposition size $M_{\text{help}}$, and normalise by $M_{\text{help}}$. With $N_{\text{help}}$ the number of subgoals the OmniBot actually inserts, this gives the closed form $\max\!\big(0,(M_{\text{help}}-N_{\text{help}})/M_{\text{help}}\big)$ for an episode that reaches the goal, and 0 for one that does not, as used in Appendix~\ref{apx:metrics}: the fraction of the oracle's work that the model's plan removes. It is bounded by CR and PR by construction, and is near 1 when the model is capable of designing near-perfect subgoal sequences. More details are presented in Appendix~\ref{apx:metrics}. 

\begin{summarybox}{Summary}\small
\benchmark is a procedurally generated gridworld, given entirely as text and used to evaluate LLMs on three tasks:
\begin{itemize}[leftmargin=*, nosep, topsep=3pt]
  \item \textbf{Predict}: can the model simulate the dynamics? 
  \item \textbf{Plan}: can it produce actions that reach a goal? 
  \item \textbf{Decompose}: can it break a mission into subgoals?
\end{itemize}
\end{summarybox}
\section{Results and Discussion}
\label{sec:results}

We evaluate seven models: four reasoning-capable models, claude-opus-4-6, claude-sonnet-4-6, gpt-5.4 and Kimi-K2.5 (open-weight), and three non-reasoning open-weight models, Llama-4-Maverick, Llama-4-Scout and Qwen2.5-72B. For all four models exposing an extended-reasoning control, extended reasoning was left at its default setting in every run; the scores below therefore characterise single-pass planning, not test-time-scaled planning. The prompt structure and querying protocol are detailed in Appendix~\ref{apx:prompts}.
\subsection{\textbf{Predict} is close to saturated for the strongest models}
\label{sec:predict-results}
\begin{table}[h]
\centering
\caption{Predict --- Success ($\uparrow$), overall.}
\label{tab:predict}
\begin{tabular}{l r r}
\toprule
Model & pass@1 & pass@3 \\
\midrule
Kimi-K2.5         & 98.0\% & 99.4\% \\
claude-opus-4-6   & 92.1\% & 98.4\% \\
claude-sonnet-4-6 & 90.7\% & 98.1\% \\
gpt-5.4           & 81.2\% & 93.4\% \\
Llama-4-Maverick  & 67.6\% & 81.2\% \\
Qwen2.5-72B       & 65.7\% & 77.8\% \\
Llama-4-Scout     & 54.9\% & 69.1\% \\
\bottomrule
\end{tabular}
\end{table}

Performance on Predict is close to saturated for the strongest models (Table~\ref{tab:predict}): the three strongest
models all exceed 90\% at pass@1 and 98\% at pass@3. The spread across models is nonetheless wide, from 98.0\% for Kimi-K2.5 down to 54.9\% for Llama-4-Scout, so the task is far from saturated for the weaker ones.

Table~\ref{tab:predict-by-length} breaks Predict down by how long a sequence the model has to simulate. Accuracy decays gradually: at 26 actions or more, Kimi-K2.5 still recovers the final state in 93\% of samples and claude-opus-4-6 in 77\%, while the Llama and Qwen models fall to between 5 and 18\%. Simulation therefore remains reliable past 26 actions for the Claude models and Kimi-K2.5, while for Llama-4-Scout and Qwen2.5-72B it has already broken down at that length.

\begin{table}[h]
\centering
\small
\caption{Predict --- pass@1 success rate (\%) by action-sequence length. Bin sizes: $n=34$, 67, 58, 38, 40 and 83 cells.}
\label{tab:predict-by-length}
\begin{tabular}{l cccccc}
\toprule
Model & 1--3 & 4--6 & 7--10 & 11--15 & 16--25 & 26+ \\
\midrule
Kimi-K2.5         & 100 & 100 & 100 & 99 & 100 & 93 \\
claude-opus-4-6   &  99 &  98 &  99 & 96 &  92 & 77 \\
claude-sonnet-4-6 & 100 &  99 &  99 & 95 &  88 & 74 \\
gpt-5.4           &  99 & 100 &  95 & 95 &  72 & 47 \\
Llama-4-Maverick  &  97 &  96 &  90 & 80 &  56 & 18 \\
Llama-4-Scout     &  94 &  88 &  69 & 65 &  40 &  5 \\
Qwen2.5-72B       &  97 &  93 &  89 & 83 &  51 & 14 \\
\bottomrule
\end{tabular}
\end{table}

For an action sequence that never meets a blocked cell or a closed door, the final state can be computed from the agent's position and orientation alone, without consulting the layout. \textbf{Predict} therefore verifies that models track the transition rules over long sequences, which is what the other two tasks presuppose step by step, but it does not on its own establish that they build a layout-aware representation. We read it as a lower bound on world-model competence, since a model that passes it can at least compute the state change of each action, which planning also requires. Sequences that cross rooms or pass through doors do depend on the layout, and the stronger models stay accurate on the levels where such sequences occur: on nine-room levels with unlocked doors Kimi-K2.5 scores 97\% and claude-opus-4-6 87\%, and on levels with a locked door 96\% and 90\% (Appendix~\ref{apx:predict-analysis}).
\begin{takeaway}
\begin{itemize}[leftmargin=*, nosep]
  \item The three strongest models simulate the dynamics accurately: $>$90\% at pass@1, $>$98\% at pass@3.
  \item Accuracy decays with sequence length, but at 26+ actions, Kimi-K2.5 and claude-opus-4-6 still reach 93\% and 77\%.
  \item For these models, misunderstanding the environment's rules is unlikely to explain planning failures.
\end{itemize}
\end{takeaway}
\subsection{\textbf{Plan} collapses at long horizons}
\label{sec:plan}
\begin{table}[h]
\centering
\caption{Plan --- Success Rate (SR, $\uparrow$), Efficiency ($\uparrow$), \texttt{ball\_distance} ($\downarrow$), overall. Efficiency is conditional on success.}
\label{tab:plan}
\begin{tabular}{l rr rr rr}
\toprule
& \multicolumn{2}{c}{SR} & \multicolumn{2}{c}{Eff} & \multicolumn{2}{c}{\texttt{ball\_dist}} \\
\cmidrule(lr){2-3}\cmidrule(lr){4-5}\cmidrule(lr){6-7}
Model & p@1 & p@3 & p@1 & p@3 & p@1 & p@3 \\
\midrule
claude-opus-4-6   & 77.2\% & 85.9\% & 0.955 & 0.986 & 2.36 & 1.18 \\
Kimi-K2.5         & 74.9\% & 95.0\% & 0.907 & 0.950 & 2.38 & 0.53 \\
claude-sonnet-4-6 & 70.9\% & 84.7\% & 0.948 & 0.973 & 2.97 & 1.22 \\
gpt-5.4           & 31.6\% & 48.1\% & 0.878 & 0.915 & 5.26 & 2.79 \\
Llama-4-Maverick  & 16.8\% & 25.3\% & 0.781 & 0.788 & 7.85 & 6.13 \\
Llama-4-Scout     &  6.9\% & 14.7\% & 0.706 & 0.721 & 10.42 & 8.12 \\
Qwen2.5-72B       &  5.1\% &  9.4\% & 0.443 & 0.571 & 10.35 & 7.88 \\
\bottomrule
\end{tabular}
\end{table}

\subsubsection{Success falls with solution length}
Completion rates on Plan are low overall (Table~\ref{tab:plan}): only three
models clear 70\% at pass@1. We found that the main factor responsible for the decreasing performance of the models is the length of the required solution (Table~\ref{tab:plan-by-length}). For every model but Llama-4-Maverick, single-attempt success is a monotone decreasing function of oracle plan length. Stratifying by path length within a fixed grid size, reproduces the collapse: claude-sonnet-4-6 falls from 97\% to 38\% across path-length terciles inside the Medium grid alone, and Llama-4-Maverick from 54\% to 0\% inside the $8\times8$ grid, whose prompts list at most eight objects; in this grid the object list varies only between five and eight entries, so longer prompts alone do not explain the collapse. Conversely, at matched obstacle count a larger grid is easier for the strong models (claude-opus-4-6: 48.3\% on Large-120 against 80.0\% on Ultra-120), since it lowers density without a proportional increase in path length. Grid size degrades completion largely because larger grids demand longer plans: the mean oracle length rises from 6.6 actions on Small to 24.4 on Ultra. Larger grids also contain fewer instances that a direct route solves, an effect analysed in Appendix~\ref{apx:lroute}. We define the \emph{planning horizon} of each model as the solution length $L_{50}$ at which single-attempt success falls to 50\%, the plan-length analogue of the horizon length $H_{0.5}$ of \citet{sinha2026longhorizon}. The planning horizon is roughly 30 actions for Kimi-K2.5, 23--25 for the Claude models, 9 for gpt-5.4, and under 7 for the Llama and Qwen models, which record almost no success beyond 15 actions. The full analysis, with per-level results, correlations and density effects, is given in Appendix~\ref{apx:plan-analysis}.

Comparing the \textbf{Plan} results with the \textbf{Predict} results of Section~\ref{sec:predict-results} shows that the strongest models cannot reliably plan in settings they can simulate. On instances whose optimal plan runs 21--30 actions, claude-opus-4-6 returns a valid plan 48.1\% of the time and Kimi-K2.5 59.6\%, and beyond 30 actions 22.2\% and 36.4\%. On action sequences of 26 or more, the same two models recover the final state 77\% and 93\% of the time. While the two tasks are evaluated on different levels, the gap is too large to attribute to that difference alone: these models simulate accurately over horizons at which they cannot reliably plan. Moreover, the mean Manhattan distance between the predicted and the true final position on Predict, averaged over all samples with correct ones counting as zero, is 0.07 to 0.51 cells for the three strongest models, against a mean \texttt{ball\_distance} of 7 to 19 cells across the seven models on the densest Ultra configuration (Ultra-180), and 7 to 9 cells for the same three models.

\begin{table}[h]
\centering
\small
\caption{Plan --- pass@1 Success Rate by oracle plan length (all grid sizes pooled), and the planning horizon $L_{50}$, obtained by linear interpolation on representative bin lengths (7, 13, 18, 25.5, 37: the midpoints of the three interior bins and a representative length for each open-ended end bin) and applied identically to every model.}
\label{tab:plan-by-length}
\begin{tabular}{l rrrrr r}
\toprule
Model & $\leq$10 & 11--15 & 16--20 & 21--30 & 31+ & $L_{50}$ \\
\midrule
Kimi-K2.5         & 88.2\% & 88.1\% & 72.1\% & 59.6\% & 36.4\% & $\approx$30 \\
claude-opus-4-6   & 99.7\% & 92.9\% & 76.9\% & 48.1\% & 22.2\% & $\approx$25 \\
claude-sonnet-4-6 & 98.3\% & 84.5\% & 58.5\% & 45.4\% & 13.1\% & $\approx$23 \\
gpt-5.4           & 61.4\% & 19.6\% & 17.7\% &  8.2\% &  6.1\% & $\approx$9 \\
Llama-4-Maverick  & 38.8\% &  8.3\% &  2.7\% &  0.0\% &  2.0\% & $<$7 \\
Llama-4-Scout     & 17.4\% &  1.8\% &  0.0\% &  0.0\% &  0.0\% & $<$7 \\
Qwen2.5-72B       & 12.4\% &  2.4\% &  0.0\% &  0.0\% &  0.0\% & $<$7 \\
\bottomrule
\end{tabular}
\end{table}

\subsubsection{Efficiency across grid sizes}
\label{sec:plan-failure}
The \textsc{LLM-BabyBench} world is simple, deterministic and fully specified. As the grid grows larger, the rules, the action set and the state representation are unchanged, and breadth-first search is able to solve every instance in polynomial time  in the grid area. Thus, we expect a system that planned by searching over the known dynamics under a fixed budget to output results that degrade gradually as the grid grows. For a reasonable grid size, the generated paths should still succeed, although the number of actions can be expected to increasingly deviate from the optimal. We observe that this is not the case in our experiments. Instead, completion collapses as the size and difficulty of the level grow, while efficiency stays stable: efficiency conditional on success stays within four points across grid sizes for the Claude models and Kimi-K2.5, and between 0.83 and 0.91 for gpt-5.4 (Table~\ref{tab:plan-by-size}). The other three models (Llama-4-Maverick, Llama-4-Scout and Qwen2.5-72B) succeed on 14, 2 and 1 of 240 Ultra samples, too few to support the ratio.\\

The responses of claude-opus-4-6 and claude-sonnet-4-6, the two models whose responses show the procedure step by step (Appendix~\ref{apx:lroute}), show one specific behaviour that is consistent with this trend: they compute the displacement to the target, propose an L-shaped corridor of two straight runs joined by a turn, scan the object list for anything lying on that corridor, shift the corridor by a row or a column when the scan reports a blocker, and commit on the first clean scan. Depending on the level, a path between the initial state and final state that can be found by this procedure is not guaranteed to exist. Some levels, for example, require moving in the opposite direction to avoid obstacles. However, when such a corridor exists, it is near-optimal by construction, since nothing shorter joins the same endpoints. Efficiency therefore stays flat across sizes, since the models only produce corridor routes, and a corridor route is either near-optimal or invalid. The chance that no corridor is clear grows with the distance to be covered, which is consistent with the rising failure rate on the larger and denser grids.
\begin{takeaway}
\begin{itemize}[leftmargin=*, nosep]
  \item Success falls with plan length. The planning horizon $L_{50}$ is $\approx$30 actions for Kimi-K2.5, 23--25 for the Claude models, 9 for gpt-5.4 and under 7 for the Llama and Qwen models.
  \item The strongest models simulate far beyond where they can plan: past 30 actions, claude-opus-4-6 and Kimi-K2.5 produce valid plans only 22\% and 36\% of the time.
  \item Successful plans stay near-optimal: the Claude models' responses show them composing L-shaped corridors.
\end{itemize}
\end{takeaway}
\subsection{\textbf{Decompose}}
\label{sec:decompose}
\begin{table}[h]
  \centering
  \small
  \caption{Decompose --- Comprehension Rate ($\uparrow$), Precision Rate
           ($\uparrow$), ACI ($\uparrow$).}
  \label{tab:decompose-overall}
  \begin{tabular}{l rr rr rr}
    \toprule
    & \multicolumn{2}{c}{CR} & \multicolumn{2}{c}{PR}
      & \multicolumn{2}{c}{ACI} \\
    \cmidrule(lr){2-3} \cmidrule(lr){4-5} \cmidrule(lr){6-7}
    Model & p@1 & p@3 & p@1 & p@3 & p@1 & p@3 \\
    \midrule
    claude-sonnet-4-6 & 88.6\% & 93.8\% & 59.9\% & 70.6\% & 0.74 & 0.83 \\
    Kimi-K2.5         & 84.9\% & 94.1\% & 72.1\% & 84.7\% & 0.79 & 0.90 \\
    gpt-5.4           & 84.1\% & 88.1\% & 51.7\% & 58.8\% & 0.66 & 0.73 \\
    Llama-4-Maverick  & 82.1\% & 88.8\% & 46.9\% & 55.6\% & 0.62 & 0.72 \\
    Llama-4-Scout     & 80.8\% & 87.2\% & 41.2\% & 48.1\% & 0.55 & 0.64 \\
    claude-opus-4-6   & 78.9\% & 94.4\% & 55.1\% & 68.8\% & 0.66 & 0.83 \\
    Qwen2.5-72B       & 76.0\% & 86.9\% & 36.1\% & 45.3\% & 0.49 & 0.61 \\
    \bottomrule
  \end{tabular}
\end{table}
\subsubsection{Comprehension and precision}
 Table~\ref{tab:decompose-overall} shows that every model achieves a high \textbf{Comprehension Rate} (76--89\% at pass@1). When an oracle
is allowed to fill in the gaps, most decompositions reach the goal, which means that the models at least recover what the mission requires and what a valid subgoal structure looks like. Except on the \texttt{PutNext} missions, where comprehension itself fails for four of the seven models (CR 3--7\%) and succeeds on at most a third of missions for two more (Table~\ref{tab:decompose-verb}), these results indicate that the models understand the task and how a subgoal decomposition works. The CR of claude-opus-4-6 is lowered by an ambiguity in how our prompt defines the argument of \texttt{GoNextToSubgoal}; restricted to the documented reading, it has the highest CR of the seven models (Appendix~\ref{apx:convention}). Yet \textbf{Precision Rate}, the success rate when the decomposition is
executed exactly as written, is far lower, 36--72\%. The models know
what to do but they cannot reliably turn it into a high-level plan that stands on its own. For stronger models, this trend is more visible in levels where the number of subgoals is higher, as displayed in Table~\ref{tab:decompose-by-count}.
\begin{table}[h]
  \centering
  \footnotesize
  \setlength{\tabcolsep}{4pt}
  \caption{Decompose --- Precision Rate by oracle subgoal count
           (pass@1 / pass@3); Comprehension Rate shown in parentheses
           beneath each value (CR pass@1 / pass@3). Bins hold 131, 109, 30, 31 and 19 cells.}
  \label{tab:decompose-by-count}
  \begin{tabular}{l ccccc}
    \toprule
    Model & 1--2 & 3--4 & 5--6 & 7--9 & 10+ \\
    \midrule
    claude-sonnet-4-6 & \prcr{92.1\% / 98.5\%}{96.2\% / 100.0\%} & \prcr{57.8\% / 75.2\%}{85.3\% / 93.6\%} & \prcr{21.1\% / 40.0\%}{88.9\% / 90.0\%} & \prcr{5.4\% / 9.7\%}{78.5\% / 80.6\%} & \prcr{0.0\% / 0.0\%}{71.9\% / 78.9\%} \\
    Kimi-K2.5         & \prcr{94.9\% / 99.2\%}{96.9\% / 100.0\%} & \prcr{70.9\% / 87.2\%}{75.5\% / 88.1\%} & \prcr{58.9\% / 86.7\%}{88.9\% / 96.7\%} & \prcr{36.6\% / 64.5\%}{74.2\% / 90.3\%} & \prcr{0.0\% / 0.0\%}{66.7\% / 89.5\%} \\
    gpt-5.4           & \prcr{90.1\% / 95.4\%}{97.7\% / 99.2\%}  & \prcr{41.6\% / 54.1\%}{74.6\% / 80.7\%} & \prcr{6.7\% / 13.3\%}{86.7\% / 93.3\%}  & \prcr{0.0\% / 0.0\%}{74.2\% / 77.4\%}  & \prcr{0.0\% / 0.0\%}{56.1\% / 63.2\%} \\
    Llama-4-Maverick  & \prcr{85.8\% / 94.7\%}{95.7\% / 99.2\%}  & \prcr{33.0\% / 45.0\%}{76.5\% / 83.5\%} & \prcr{5.6\% / 16.7\%}{81.1\% / 93.3\%}  & \prcr{0.0\% / 0.0\%}{63.4\% / 77.4\%}  & \prcr{0.0\% / 0.0\%}{52.6\% / 57.9\%} \\
    Llama-4-Scout     & \prcr{77.9\% / 87.0\%}{95.7\% / 100.0\%} & \prcr{27.2\% / 35.8\%}{72.2\% / 80.7\%} & \prcr{1.1\% / 3.3\%}{83.3\% / 90.0\%}   & \prcr{0.0\% / 0.0\%}{60.2\% / 71.0\%}  & \prcr{0.0\% / 0.0\%}{57.9\% / 57.9\%} \\
    claude-opus-4-6   & \prcr{77.6\% / 87.8\%}{84.7\% / 93.1\%}  & \prcr{59.3\% / 80.7\%}{79.5\% / 97.2\%} & \prcr{18.9\% / 33.3\%}{75.6\% / 96.7\%} & \prcr{14.0\% / 22.6\%}{60.2\% / 87.1\%} & \prcr{0.0\% / 0.0\%}{70.2\% / 94.7\%} \\
    Qwen2.5-72B       & \prcr{68.7\% / 83.2\%}{91.9\% / 99.2\%}  & \prcr{23.2\% / 32.1\%}{66.7\% / 78.9\%} & \prcr{1.1\% / 3.3\%}{73.3\% / 90.0\%}   & \prcr{0.0\% / 0.0\%}{59.1\% / 77.4\%}  & \prcr{0.0\% / 0.0\%}{52.6\% / 57.9\%} \\
    \bottomrule
  \end{tabular}
\end{table}
Across all models, standalone decomposition performance is high for very short missions (1–2 subgoals), where it suffices to translate the task from Baby Language into a succession of subgoals, but collapses immediately for slightly longer missions and vanishes for tasks that require ten or more subgoals (PR = 0.0\% across the board at 10+). For example, claude-sonnet-4-6 goes from 92\% on missions that need 1–2 subgoals to 58\% and 21\% for missions that need 3–4 and 5–6 subgoals, respectively (see Table~\ref{tab:decompose-by-count}). PR is a demanding metric by construction, as it forbids any runtime insertion. Hence, the model must anticipate every step that the bot would otherwise supply reactively, including subgoals that depend on the bot's own conventions, such as the \texttt{Drop} it appends whenever a mission involves picking something up (Appendix~\ref{apx:drop}). Therefore, it may be too restrictive to judge models solely by the PR metric. A subgoal decomposition that is not self-sufficient can still be used as a fast first guess before a slow verifier validates and completes it.

\subsubsection{Assistance to a solver}
The Assistance Curve Integral (ACI) measures the practical worth of such a first guess: how much of a downstream solver's work it removes. Several works propose using LLMs as approximate solvers whose output classical deterministic algorithms verify or complete. The performance of such models when incorporated into complex mechanisms is hard to assess in general. ACI provides a simple numerical measure of this performance in \benchmark.

Overall (Table~\ref{tab:decompose-overall}), ACI ranges from 0.49 to 0.79 at pass@1: the best model (Kimi-K2.5, 0.79) hands a solver a plan that is most of the way there, while the weakest (Qwen2.5-72B, 0.49) leaves the solver with roughly half the work. However, a cheap approximator is most needed when the horizon is long enough that the search space is too large. Hence, ACI is most informative for long tasks, since this is where a deterministic solver grows expensive and we may want an automated model to supply a first guess that reduces the solver's workload. Unfortunately, ACI suffers from the same decline in performance with the number of subgoals needed.

At pass@1, ACI decays monotonically with horizon for every model (Table~\ref{tab:decompose-aci-by-count}; claude-opus-4-6 falls from 0.81 at 1--2 subgoals to 0.32 at 10+, Kimi-K2.5 from 0.96 to 0.38), so the LLM's draft is worth least to the solver on the long tasks where automation is most wanted. For short-horizon decomposition, an LLM removes most of a solver's work and an LLM-plus-solver pipeline is attractive, but for long-horizon decomposition, today's models do not yet provide a draft worth refining, and the solver must still do most of the search.

\begin{table}[ht]
  \centering
  \small
  \caption{Decompose --- ACI by oracle subgoal count (pass@1 / pass@3).}
  \label{tab:decompose-aci-by-count}
  \begin{tabular}{l ccccc}
    \toprule
    Model & 1--2 & 3--4 & 5--6 & 7--9 & 10+ \\
    \midrule
    claude-sonnet-4-6 & 0.94 / 0.99 & 0.72 / 0.85 & 0.56 / 0.65 & 0.44 / 0.53 & 0.26 / 0.36 \\
    Kimi-K2.5         & 0.96 / 1.00 & 0.73 / 0.88 & 0.72 / 0.93 & 0.57 / 0.81 & 0.38 / 0.54 \\
    gpt-5.4           & 0.93 / 0.97 & 0.57 / 0.67 & 0.43 / 0.55 & 0.35 / 0.42 & 0.21 / 0.28 \\
    Llama-4-Maverick  & 0.90 / 0.96 & 0.52 / 0.66 & 0.40 / 0.56 & 0.25 / 0.38 & 0.18 / 0.25 \\
    Llama-4-Scout     & 0.85 / 0.93 & 0.44 / 0.55 & 0.29 / 0.35 & 0.20 / 0.28 & 0.16 / 0.22 \\
    claude-opus-4-6   & 0.81 / 0.90 & 0.69 / 0.90 & 0.46 / 0.69 & 0.36 / 0.55 & 0.32 / 0.52 \\
    Qwen2.5-72B       & 0.76 / 0.89 & 0.38 / 0.51 & 0.24 / 0.35 & 0.18 / 0.28 & 0.12 / 0.18 \\
    \bottomrule
  \end{tabular}
\end{table}

\subsubsection{Omissions in long decompositions}

The results show that models generally understand what the mission asks them to do, but fail to plan the world interactions required to execute it. Comprehension rate stays high even for missions of 10+ subgoals whereas ACI and PR collapse at that length. This is not primarily a problem of ordering: the subgoals models emit are generally in a workable order, and CR stays between 53 and 72\% on the same instances (Table~\ref{tab:decompose-by-count}). This behaviour is due to the models systematically omitting the execution machinery needed to accomplish tasks given, especially actions such as opening doors and clearing blocked paths (Table~\ref{tab:decompose-audit}). For example, in a BossLevel mission asking it to ``pick up a green key after opening a red door and going to the blue door'', gpt-5.4 includes an \texttt{OpenSubgoal} for the red door but none for the other doors the route goes through (Appendix~\ref{apx:decompose-audit}).

\subsubsection{Which missions the oracle can plan up front}
Some levels in \benchmark necessitate more careful planning than others. Suppose a level requires picking up a ball, but the path towards that object is blocked by a closed red door. The model must anticipate this obstacle and plan the necessary subgoals in advance: first picking up a red key and then using it to open the door before proceeding to the target object. OmniBot uses another strategy: since it operates directly in the environment, it starts with only the main subgoal (Pick up ball) and, when it reaches the closed red door, improvises the necessary subgoals to solve the level.\\
Whether the level is static (OmniBot does not improvise) or not, and how many subgoals OmniBot inserts at runtime after crafting its initial plan, can be seen as another proxy for level difficulty. On the 177 cells where the oracle never improvises, PR is between 64 and 88\% at pass@1 for every model. On the 69 cells where it has to add three or more subgoals of its own, PR is between 0 and 12\% for six models and 44\% for Kimi-K2.5, while CR stays between 70 and 87\% throughout (Appendix~\ref{apx:decompose-analysis}). Whether a mission can be planned in full by the oracle before execution begins is a strong predictor of PR, although for every model at least one of subgoal count, path length and room count correlates with PR more strongly (Table~\ref{tab:decompose-correlations}), and models fail predominantly in non-static levels. Another display of this behaviour is how models perform on one-room levels. In levels of this type, since there are no doors, OmniBot can always solve the task only relying on its initial plan. In those levels, CR and PR coincide for every model but Qwen2.5-72B, so PR and ACI separate the models mainly on multi-room levels, where PR ranges from 12\% (Qwen2.5-72B) to 65\% (Kimi-K2.5) (Table~\ref{tab:decompose-structure}). 
\begin{takeaway}
\begin{itemize}[leftmargin=*, nosep]
  \item Models understand the missions (CR 76--89\%) but often fail to write self-sufficient plans (PR 36--72\%). PR is 0\% for every model at 10+ subgoals.
  \item Long plans fail by omission: they leave out door openings and path clearing, mostly in missions where the oracle must improvise at runtime.
  \item LLM drafts help a solver mostly on short missions.
\end{itemize}
\end{takeaway}
\section{Limitations}
\label{sec:limitations}
\benchmark removes the factors that
usually accompany planning in interactive benchmarks (perception, partial observability, instruction ambiguity,
retrieval, commonsense understanding of the effects of actions, unknown dynamics), so the scores we report are not confounded by them. Some non-planning factors remain: the Precision Rate requires models to reproduce an executor convention (Appendix~\ref{apx:drop}), our definition of the \texttt{GoNextToSubgoal} argument admits two readings (Appendix~\ref{apx:convention}), and on the largest grids a plan must be checked against a list of up to 180 objects (Appendix~\ref{apx:lroute}). With respect to these removed factors, the performance shown in \benchmark should be seen as an upper bound on the planning performance of grounded LLMs. In practice an agent must plan while it is also seeing, remembering, disambiguating and forming beliefs in a partially observable world, and there its performance would be worse than what we report here. We do not measure how much each removed factor costs a planner in practice.

The four models that expose an extended-reasoning control were run with its default setting, so our results characterise single-pass planning; test-time scaling may extend the horizons we report. \textbf{Predict} and \textbf{Plan} are also evaluated on different levels, so the comparison between simulation and planning contrasts trends rather than matched instances.

Another limitation is that BabyAI and MiniGrid have been open-source for years and the levels we use carry their original names, so we cannot rule out that models have seen descriptions of these dynamics, or expert trajectories, during training. Memorisation is a limited threat here, since instances are procedurally generated from seeds and scored by execution. The low \textbf{Plan} and \textbf{Decompose} scores at long horizons also suggest that any such exposure does not translate into strong planning performance.

\section{Conclusion}
We propose \benchmark, a fully observable textual gridworld in which perception, instruction ambiguity, world knowledge, unknown dynamics and retrieval are controlled by construction (Section~\ref{sec:limitations} notes the residual non-planning factors), together with the PPD suite and an evaluation harness in which every answer is checked by execution. Relying on this environment, we show that, queried zero-shot with extended reasoning at its default setting, the strongest models cannot reliably plan in environments they simulate accurately.

In \textbf{Plan}, single-attempt success collapses as the required plan grows longer. In \textbf{Decompose}, models still understand what a mission requires well beyond the length at which they can write a complete subgoal sequence for it, and when a deterministic bot completes their drafts, the drafts save it little work on long missions. In both tasks, the reliability of plan synthesis decreases with the number of sequential decisions a plan requires, counted in primitive actions for \textbf{Plan} and in subgoals for \textbf{Decompose}. For the strongest models, the main cause of failure is therefore plan synthesis: they simulate the environment well enough to check a plan they are given. Because our setting controls the other factors listed in Section~\ref{sec:omnibot}, these scores are an upper bound on what the same models, queried under the same protocol, can be expected to achieve in grounded applications. We hope \benchmark can serve as a controlled testbed for studying what would let LLMs plan better.

\impact{\benchmark is a synthetic, procedurally generated gridworld. It contains no personal data and involves no human subjects. Its intended use is to measure whether language models can plan when perception, instruction ambiguity, world knowledge, unknown dynamics and retrieval are controlled by construction. The main risk of misuse is to read a high score on it as evidence that a model can plan in real-world settings. The benchmark controls the factors that make real-world planning harder, so, for a given model and protocol, its scores should be read as an upper bound on real-world planning performance, not on what the model can reach with more test-time computation. Better planning in language-model agents would make autonomous agents more capable in settings where their actions have consequences; the benchmark is designed to measure this capability, not to train it, although its released expert action sequences could be used as training data, which would also contaminate the public seeds. On the positive side, a controlled measure of where and why planning fails can help practitioners decide when an agent's plans need an external verifier before deployment.}

\acks{This work was supported by Mohamed bin Zayed University of Artificial Intelligence (MBZUAI). The authors declare no competing interests.}

\vskip 0.2in
\bibliography{ref_dmlr}

@inproceedings{planbench,
  title={{PlanBench}: An Extensible Benchmark for Evaluating Large Language Models on Planning and Reasoning about Change},
  author={Valmeekam, Karthik and Marquez, Matthew and Olmo, Alberto and Sreedharan, Sarath and Kambhampati, Subbarao},
  booktitle={Advances in Neural Information Processing Systems (NeurIPS) 36, Datasets and Benchmarks Track},
  pages={38975--38987},
  year={2023}
}

@inproceedings{valmeekam2023planning,
  title={On the Planning Abilities of Large Language Models -- A Critical Investigation},
  pages={75993--76005},
  author={Valmeekam, Karthik and Marquez, Matthew and Sreedharan, Sarath and Kambhampati, Subbarao},
  booktitle={Advances in Neural Information Processing Systems (NeurIPS) 36},
  year={2023}
}

@inproceedings{savva2019habitat,
  title={Habitat: A platform for embodied {AI} research},
  author={Savva, Manolis and Kadian, Abhishek and Maksymets, Oleksandr and Zhao, Yili and Wijmans, Erik and Jain, Bhavana and Straub, Julian and Liu, Jia and Koltun, Vladlen and Malik, Jitendra and others},
  booktitle={Proceedings of the IEEE/CVF international conference on computer vision},
  pages={9339--9347},
  year={2019}
}

@inproceedings{grounding_online_rl,
author = {Carta, Thomas and Romac, Cl\'{e}ment and Wolf, Thomas and Lamprier, Sylvain and Sigaud, Olivier and Oudeyer, Pierre-Yves},
title = {Grounding large language models in interactive environments with online reinforcement learning},
year = {2023},
publisher = {JMLR.org},
booktitle = {Proceedings of the 40th International Conference on Machine Learning},
articleno = {150},
numpages = {38},
location = {Honolulu, Hawaii, USA},
series = {ICML'23}
}

@inproceedings{liang2023codeaspolicies,
  title={Code as policies: Language model programs for embodied control},
  author={Liang, Jacky and Huang, Wenlong and Xia, Fei and Xu, Peng and Hausman, Karol and Ichter, Brian and Florence, Pete and Zeng, Andy},
  booktitle={2023 IEEE International Conference on Robotics and Automation (ICRA)},
  pages={9493--9500},
  year={2023},
  organization={IEEE}
}

@inproceedings{MinigridMiniworld23,
  author       = {Maxime Chevalier{-}Boisvert and Bolun Dai and Mark Towers and Rodrigo Perez{-}Vicente and Lucas Willems and Salem Lahlou and Suman Pal and Pablo Samuel Castro and Jordan Terry},
  title        = {{Minigrid} {\&} {Miniworld}: Modular {\&} Customizable Reinforcement Learning Environments for Goal-Oriented Tasks},
  booktitle    = {Advances in Neural Information Processing Systems 36, Datasets and Benchmarks Track},
  month        = {December},
  year         = {2023},
}

@inproceedings{fan2022minedojo,
  title     = {{MineDojo}: Building Open-Ended Embodied Agents with Internet-Scale Knowledge},
  author    = {Linxi Fan and Guanzhi Wang and Yunfan Jiang and Ajay Mandlekar and Yuncong Yang and Haoyi Zhu and Andrew Tang and De-An Huang and Yuke Zhu and Anima Anandkumar},
  booktitle = {Thirty-sixth Conference on Neural Information Processing Systems Datasets and Benchmarks Track},
  year      = {2022},
  url       = {https://openreview.net/forum?id=rc8o_j8I8PX}
}

@inproceedings{puerto-etal-2024-code,
    title = "Code Prompting Elicits Conditional Reasoning Abilities in {T}ext+{C}ode {LLM}s",
    author = "Puerto, Haritz  and
      Tutek, Martin  and
      Aditya, Somak  and
      Zhu, Xiaodan  and
      Gurevych, Iryna",
    editor = "Al-Onaizan, Yaser  and
      Bansal, Mohit  and
      Chen, Yun-Nung",
    booktitle = "Proceedings of the 2024 Conference on Empirical Methods in Natural Language Processing",
    month = nov,
    year = "2024",
    address = "Miami, Florida, USA",
    pages = "11234--11258",
    publisher = "Association for Computational Linguistics",
}

@misc{he2024doespromptformattingimpact,
      title={Does Prompt Formatting Have Any Impact on {LLM} Performance?}, 
      author={Jia He and Mukund Rungta and David Koleczek and Arshdeep Sekhon and Franklin X Wang and Sadid Hasan},
      year={2024},
      eprint={2411.10541},
      archivePrefix={arXiv},
      primaryClass={cs.CL},
      url={https://arxiv.org/abs/2411.10541}, 
}

@inproceedings{babyai,
  title     = {{BabyAI}: A Platform to Study the Sample Efficiency of Grounded
               Language Learning},
  author    = {Chevalier-Boisvert, Maxime and Bahdanau, Dzmitry and
               Lahlou, Salem and Willems, Lucas and Saharia, Chitwan and
               Nguyen, Thien Huu and Bengio, Yoshua},
  booktitle = {International Conference on Learning Representations (ICLR)},
  year      = {2019},
  note      = {arXiv:1810.08272}
}

@inproceedings{alfred,
  title     = {{ALFRED}: A Benchmark for Interpreting Grounded Instructions
               for Everyday Tasks},
  author    = {Shridhar, Mohit and Thomason, Jesse and Gordon, Daniel and
               Bisk, Yonatan and Han, Winson and Mottaghi, Roozbeh and
               Zettlemoyer, Luke and Fox, Dieter},
  booktitle = {IEEE/CVF Conference on Computer Vision and Pattern Recognition
               (CVPR)},
  pages     = {10737--10746},
  year      = {2020}
}

@inproceedings{ALFWorld20,
  title     = {{ALFWorld}: Aligning Text and Embodied Environments for
               Interactive Learning},
  author    = {Shridhar, Mohit and Yuan, Xingdi and C{\^o}t{\'e},
               Marc-Alexandre and Bisk, Yonatan and Trischler, Adam and
               Hausknecht, Matthew},
  booktitle = {International Conference on Learning Representations (ICLR)},
  year      = {2021}
}

@inproceedings{cote2018textworld,
  title     = {{TextWorld}: A Learning Environment for Text-Based Games},
  author    = {C{\^o}t{\'e}, Marc-Alexandre and K{\'a}d{\'a}r, {\'A}kos and
               Yuan, Xingdi and Kybartas, Ben and Barnes, Tavian and
               Fine, Emery and Moore, James and Hausknecht, Matthew and
               El Asri, Layla and Adada, Mahmoud and Tay, Wendy and
               Trischler, Adam},
  booktitle = {Workshop on Computer Games at IJCAI},
  pages     = {41--75},
  year      = {2018}
}

@inproceedings{wang2022scienceworld,
  title     = {{ScienceWorld}: Is Your Agent Smarter than a 5th Grader?},
  author    = {Wang, Ruoyao and Jansen, Peter and C{\^o}t{\'e},
               Marc-Alexandre and Ammanabrolu, Prithviraj},
  booktitle = {Conference on Empirical Methods in Natural Language Processing
               (EMNLP)},
  year      = {2022}
}

@article{wang2023voyager,
  title   = {Voyager: An Open-Ended Embodied Agent with Large Language Models},
  author  = {Wang, Guanzhi and Xie, Yuqi and Jiang, Yunfan and
             Mandlekar, Ajay and Xiao, Chaowei and Zhu, Yuke and
             Fan, Linxi and Anandkumar, Anima},
  journal = {Transactions on Machine Learning Research (TMLR)},
  year    = {2024}
}

@inproceedings{paglieri2025balrog,
  title     = {{BALROG}: Benchmarking Agentic {LLM} and {VLM} Reasoning on
               Games},
  author    = {Paglieri, Davide and Cupia{\l}, Bart{\l}omiej and
               Coward, Samuel and Piterbarg, Ulyana and Wolczyk, Maciej and
               Khan, Akbir and Pignatelli, Eduardo and Kuci{\'n}ski, {\L}ukasz
               and Pinto, Lerrel and Fergus, Rob and Foerster, Jakob N. and
               Parker-Holder, Jack and Rockt{\"a}schel, Tim},
  booktitle = {International Conference on Learning Representations (ICLR)},
  year      = {2025}
}

@inproceedings{wu2024smartplay,
  title     = {{SmartPlay}: A Benchmark for {LLMs} as Intelligent Agents},
  author    = {Wu, Yue and Tang, Xuan and Mitchell, Tom M. and Li, Yuanzhi},
  booktitle = {International Conference on Learning Representations (ICLR)},
  year      = {2024}
}

@inproceedings{liu2024agentbench,
  title     = {{AgentBench}: Evaluating {LLMs} as Agents},
  author    = {Liu, Xiao and Yu, Hao and Zhang, Hanchen and Xu, Yifan and
               Lei, Xuanyu and Lai, Hanyu and Gu, Yu and Ding, Hangliang and
               Men, Kaiwen and Yang, Kejuan and Zhang, Shudan and
               Deng, Xiang and Zeng, Aohan and Du, Zhengxiao and
               Zhang, Chenhui and Shen, Sheng and Zhang, Tianjun and
               Su, Yu and Sun, Huan and Huang, Minlie and Dong, Yuxiao and
               Tang, Jie},
  booktitle = {International Conference on Learning Representations (ICLR)},
  year      = {2024}
}

@inproceedings{samvelyan2021minihack,
  title     = {{MiniHack} the Planet: A Sandbox for Open-Ended Reinforcement
               Learning Research},
  author    = {Samvelyan, Mikayel and Kirk, Robert and Kurin, Vitaly and
               Parker-Holder, Jack and Jiang, Minqi and Hambro, Eric and
               Petroni, Fabio and K{\"u}ttler, Heinrich and Grefenstette, Edward
               and Rockt{\"a}schel, Tim},
  booktitle = {Proceedings of the Neural Information Processing Systems Track on
               Datasets and Benchmarks},
  year      = {2021}
}

@inproceedings{kuttler2020nethack,
  title     = {The {NetHack} Learning Environment},
  author    = {K{\"u}ttler, Heinrich and Nardelli, Nantas and Miller, Alexander H.
               and Raileanu, Roberta and Selvatici, Marco and
               Grefenstette, Edward and Rockt{\"a}schel, Tim},
  booktitle = {Advances in Neural Information Processing Systems (NeurIPS)},
  year      = {2020}
}

@inproceedings{hafner2022crafter,
  title     = {Benchmarking the Spectrum of Agent Capabilities},
  author    = {Hafner, Danijar},
  booktitle = {International Conference on Learning Representations (ICLR)},
  year      = {2022}
}

@article{valmeekam2025o1,
  title   = {A Systematic Evaluation of the Planning and Scheduling Abilities
             of the Reasoning Model {o1}},
  author  = {Valmeekam, Karthik and Stechly, Kaya and Gundawar, Atharva and
             Kambhampati, Subbarao},
  journal = {Transactions on Machine Learning Research (TMLR)},
  year    = {2025}
}

@inproceedings{kambhampati2024llmmodulo,
  title     = {Position: {LLMs} Can't Plan, But Can Help Planning in
               {LLM-Modulo} Frameworks},
  author    = {Kambhampati, Subbarao and Valmeekam, Karthik and Guan, Lin and
               Verma, Mudit and Stechly, Kaya and Bhambri, Siddhant and
               Saldyt, Lucas and Murthy, Anil B},
  booktitle = {International Conference on Machine Learning (ICML)},
  series    = {Proceedings of Machine Learning Research},
  volume    = {235},
  pages     = {22895--22907},
  year      = {2024}
}

@article{naturalplan,
  title   = {{NATURAL PLAN}: Benchmarking {LLMs} on Natural Language Planning},
  author  = {Zheng, Huaixiu Steven and Mishra, Swaroop and Zhang, Hugh and
             Chen, Xinyun and Chen, Minmin and Nova, Azade and Hou, Le and
             Cheng, Heng-Tze and Le, Quoc V. and Chi, Ed H. and Zhou, Denny},
  journal = {arXiv preprint arXiv:2406.04520},
  year    = {2024}
}

@inproceedings{xie2024travelplanner,
  title     = {{TravelPlanner}: A Benchmark for Real-World Planning with
               Language Agents},
  author    = {Xie, Jian and Zhang, Kai and Chen, Jiangjie and Zhu, Tinghui and
               Lou, Renze and Tian, Yuandong and Xiao, Yanghua and Su, Yu},
  booktitle = {International Conference on Machine Learning (ICML)},
  year      = {2024}
}

@article{aghzal2023ppnl,
  title   = {Can Large Language Models be Good Path Planners? {A} Benchmark and
             Investigation on Spatial-Temporal Reasoning},
  author  = {Aghzal, Mohamed and Plaku, Erion and Yao, Ziyu},
  journal = {ICLR 2024 Workshop on Large Language Model (LLM) Agents; arXiv preprint arXiv:2310.03249},
  year    = {2023}
}

@inproceedings{aghzal2024lookfurther,
  title={Look further ahead: Testing the Limits of {GPT-4} in Path Planning},
  author={Aghzal, Mohamed and Plaku, Erion and Yao, Ziyu},
  booktitle={2024 IEEE 20th International Conference on Automation Science and Engineering (CASE)},
  pages={1020--1027},
  year={2024},
  organization={IEEE}
}

@inproceedings{momennejad2023cogeval,
  title     = {Evaluating Cognitive Maps and Planning in Large Language Models
               with {CogEval}},
  author    = {Momennejad, Ida and Hasanbeig, Hosein and
               Vieira Frujeri, Felipe and Sharma, Hiteshi and Jojic, Nebojsa
               and Palangi, Hamid and Ness, Robert and Larson, Jonathan},
  booktitle = {Advances in Neural Information Processing Systems (NeurIPS)},
  volume    = {36},
  year      = {2023}
}

@inproceedings{li2024embodiedagentinterface,
  title     = {Embodied Agent Interface: Benchmarking {LLMs} for Embodied
               Decision Making},
  author    = {Li, Manling and Zhao, Shiyu and Wang, Qineng and Wang, Kangrui
               and Zhou, Yu and Srivastava, Sanjana and Gokmen, Cem and
               Lee, Tony and Li, Li Erran and Zhang, Ruohan and Liu, Weiyu and
               Liang, Percy and Fei-Fei, Li and Mao, Jiayuan and Wu, Jiajun},
  booktitle = {Advances in Neural Information Processing Systems (NeurIPS)
               Datasets and Benchmarks Track},
  year      = {2024}
}

@inproceedings{vafa2024worldmodel,
  title     = {Evaluating the World Model Implicit in a Generative Model},
  author    = {Vafa, Keyon and Chen, Justin Y. and Rambachan, Ashesh and
               Kleinberg, Jon and Mullainathan, Sendhil},
  booktitle = {Advances in Neural Information Processing Systems (NeurIPS)},
  volume    = {37},
  year      = {2024}
}

@inproceedings{li2023othello,
  title     = {Emergent World Representations: Exploring a Sequence Model
               Trained on a Synthetic Task},
  author    = {Li, Kenneth and Hopkins, Aspen K. and Bau, David and
               Vi{\'e}gas, Fernanda and Pfister, Hanspeter and
               Wattenberg, Martin},
  booktitle = {International Conference on Learning Representations (ICLR)},
  year      = {2023}
}

@inproceedings{gurnee2024space,
  title     = {Language Models Represent Space and Time},
  author    = {Gurnee, Wes and Tegmark, Max},
  booktitle = {International Conference on Learning Representations (ICLR)},
  year      = {2024}
}

@inproceedings{hao2023rap,
  title     = {Reasoning with Language Model is Planning with World Model},
  author    = {Hao, Shibo and Gu, Yi and Ma, Haodi and Hong, Joshua Jiahua and
               Wang, Zhen and Wang, Daisy Zhe and Hu, Zhiting},
  booktitle = {Conference on Empirical Methods in Natural Language Processing
               (EMNLP)},
  year      = {2023}
}

@inproceedings{shojaee2025illusion,
  title   = {The Illusion of Thinking: Understanding the Strengths and
             Limitations of Reasoning Models via the Lens of Problem
             Complexity},
  author  = {Shojaee, Parshin and Mirzadeh, Iman and Alizadeh, Keivan and
             Horton, Maxwell and Bengio, Samy and Farajtabar, Mehrdad},
  booktitle = {Advances in Neural Information Processing Systems (NeurIPS)},
  year    = {2025}
}

@inproceedings{sinha2026longhorizon,
  title     = {The Illusion of Diminishing Returns: Measuring Long Horizon
               Execution in {LLMs}},
  author    = {Sinha, Akshit and Arun, Arvindh and Goel, Shashwat and
               Staab, Steffen and Geiping, Jonas},
  booktitle = {International Conference on Learning Representations (ICLR)},
  year      = {2026},
  note      = {arXiv:2509.09677}
}

@inproceedings{ahn2022saycan,
  title     = {Do As {I} Can, Not As {I} Say: Grounding Language in Robotic
               Affordances},
  author    = {Ahn, Michael and Brohan, Anthony and Brown, Noah and
               Chebotar, Yevgen and Cortes, Omar and David, Byron and
               Finn, Chelsea and Fu, Chuyuan and Gopalakrishnan, Keerthana and
               Hausman, Karol and Herzog, Alex and Ho, Daniel and
               Hsu, Jasmine and Ibarz, Julian and Ichter, Brian and
               Irpan, Alex and Jang, Eric and Ruano, Rosario Jauregui and
               Jeffrey, Kyle and Jesmonth, Sally and Joshi, Nikhil and
               Julian, Ryan and Kalashnikov, Dmitry and Kuang, Yuheng and
               Lee, Kuang-Huei and Levine, Sergey and Lu, Yao and
               Luu, Linda and Parada, Carolina and Pastor, Peter and
               Quiambao, Jornell and Rao, Kanishka and Rettinghouse, Jarek and
               Reyes, Diego and Sermanet, Pierre and Sievers, Nicolas and
               Tan, Clayton and Toshev, Alexander and Vanhoucke, Vincent and
               Xia, Fei and Xiao, Ted and Xu, Peng and Xu, Sichun and
               Yan, Mengyuan and Zeng, Andy},
  booktitle = {Conference on Robot Learning (CoRL)},
  year      = {2022}
}

@inproceedings{huang2022zeroshotplanners,
  title     = {Language Models as Zero-Shot Planners: Extracting Actionable
               Knowledge for Embodied Agents},
  author    = {Huang, Wenlong and Abbeel, Pieter and Pathak, Deepak and
               Mordatch, Igor},
  booktitle = {International Conference on Machine Learning (ICML)},
  year      = {2022}
}

@inproceedings{huang2022innermonologue,
  title     = {Inner Monologue: Embodied Reasoning through Planning with
               Language Models},
  author    = {Huang, Wenlong and Xia, Fei and Xiao, Ted and Chan, Harris and
               Liang, Jacky and Florence, Pete and Zeng, Andy and
               Tompson, Jonathan and Mordatch, Igor and Chebotar, Yevgen and
               Sermanet, Pierre and Brown, Noah and Jackson, Tomas and
               Luu, Linda and Levine, Sergey and Hausman, Karol and
               Ichter, Brian},
  booktitle = {Conference on Robot Learning (CoRL)},
  year      = {2022}
}

@inproceedings{singh2023progprompt,
  title     = {{ProgPrompt}: Generating Situated Robot Task Plans using Large
               Language Models},
  author    = {Singh, Ishika and Blukis, Valts and Mousavian, Arsalan and
               Goyal, Ankit and Xu, Danfei and Tremblay, Jonathan and
               Fox, Dieter and Thomason, Jesse and Garg, Animesh},
  booktitle = {IEEE International Conference on Robotics and Automation (ICRA)},
  year      = {2023}
}

@article{liu2023llmp,
  title   = {{LLM+P}: Empowering Large Language Models with Optimal Planning
             Proficiency},
  author  = {Liu, Bo and Jiang, Yuqian and Zhang, Xiaohan and Liu, Qiang and
             Zhang, Shiqi and Biswas, Joydeep and Stone, Peter},
  journal = {arXiv preprint arXiv:2304.11477},
  year    = {2023}
}

@inproceedings{song2023llmplanner,
  title     = {{LLM-Planner}: Few-Shot Grounded Planning for Embodied Agents
               with Large Language Models},
  author    = {Song, Chan Hee and Wu, Jiaman and Washington, Clayton and
               Sadler, Brian M. and Chao, Wei-Lun and Su, Yu},
  booktitle = {IEEE/CVF International Conference on Computer Vision (ICCV)},
  year      = {2023}
}

@inproceedings{lehnert2024searchformer,
  title   = {Beyond {A*}: Better Planning with Transformers via Search
             Dynamics Bootstrapping},
  author  = {Lehnert, Lucas and Sukhbaatar, Sainbayar and Su, DiJia and
             Zheng, Qinqing and McVay, Paul and Rabbat, Michael and
             Tian, Yuandong},
  booktitle = {Conference on Language Modeling (COLM)},
  year    = {2024}
}

@inproceedings{gandhi2024streamofsearch,
  title   = {Stream of Search ({SoS}): Learning to Search in Language},
  author  = {Gandhi, Kanishk and Lee, Denise and Grand, Gabriel and
             Liu, Muxin and Cheng, Winson and Sharma, Archit and
             Goodman, Noah D.},
  booktitle = {Conference on Language Modeling (COLM)},
  year    = {2024}
}

@article{sutton1999options,
  title   = {Between {MDPs} and semi-{MDPs}: A Framework for Temporal
             Abstraction in Reinforcement Learning},
  author  = {Sutton, Richard S. and Precup, Doina and Singh, Satinder},
  journal = {Artificial Intelligence},
  volume  = {112},
  number  = {1-2},
  pages   = {181--211},
  year    = {1999}
}

@inproceedings{
ye2025justice,
title={Justice or Prejudice? {Q}uantifying Biases in {LLM}-as-a-Judge},
author={Jiayi Ye and Yanbo Wang and Yue Huang and Dongping Chen and Qihui Zhang and Nuno Moniz and Tian Gao and Werner Geyer and Chao Huang and Pin-Yu Chen and Nitesh V Chawla and Xiangliang Zhang},
booktitle={The Thirteenth International Conference on Learning Representations},
year={2025},

}
\appendix
\section{Experimental Details and Further Analyses}

\subsection{Experimental details}
\subsubsection{The 16 BabyAI Levels}
\label{apx:levels}

The BabyAI paper defines 19 levels. We keep 16 of them, listed with their required skills in Table~\ref{tab:babyailevels}: GoToObj, GoToRedBallGrey, GoToRedBall, GoToLocal, PutNextLocal, PickupLoc, GoToObjMaze, GoTo, Pickup, UnblockPickup, Open, Synth, SynthLoc, GoToSeq, SynthSeq and BossLevel. They are grouped into four difficulty bands, from single-room navigation with no distractors up to BossLevel, which combines maze navigation, unblocking, unlocking and multi-clause instructions.

\begin{figure}[h]
  \centering
  \includegraphics[width=0.45\textwidth]{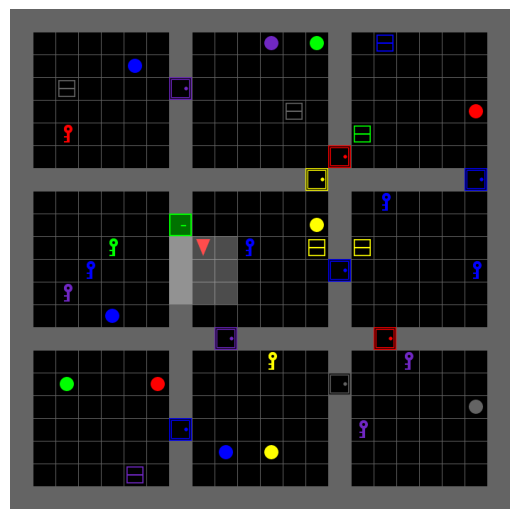}
  \caption{An Unlock instance (seed 63) with the mission ``open the green door''. The only green key sits behind the locked green door it is needed to open. Instances of this kind are excluded from \benchmark.}
  \label{fig:unlock63}
\end{figure}

The three excluded levels are dropped because they can generate instances that no agent can solve, ours included. Figure~\ref{fig:unlock63} shows one: an Unlock instance whose mission is to open the green door, where the only green key has been placed inside a room locked by that same green door. Such an instance cannot be solved under any observability, and an unsolvable instance would corrupt every metric we report. Filtering them also makes the OmniBot's $100\%$ solve rate meaningful, since the rate is measured on a pool from which the degenerate configurations have been removed.

\begin{table}[h]

\centering
\caption{The 16 BabyAI levels retained in \benchmark, grouped by difficulty, with the skills each level exercises under the taxonomy of \citet{babyai}.}
\label{tab:babyailevels}
\resizebox{\textwidth}{!}{\begin{tabular}{>{\scriptsize}p{3.85cm}|cc|ccccccc|ccc|cccc}
\hline
\multirow{2}{*}{\textbf{Skills}} &
\multicolumn{2}{c|}{\textbf{Easy}} &
\multicolumn{7}{c|}{\textbf{Medium}} &
\multicolumn{3}{c|}{\textbf{Hard}} &
\multicolumn{4}{c}{\textbf{Very Hard}} \\
\cline{2-17}

& \rotatebox{90}{GoToObj} &
\rotatebox{90}{GoToRedBallGrey} &
\rotatebox{90}{GoToRedBall} &
\rotatebox{90}{GoToLocal} &
\rotatebox{90}{PutNextLocal} &
\rotatebox{90}{PickupLoc} &
\rotatebox{90}{GoToObjMaze} &
\rotatebox{90}{GoTo} &
\rotatebox{90}{Pickup} &
\rotatebox{90}{UnblockPickup} &
\rotatebox{90}{Open} &
\rotatebox{90}{Synth} &
\rotatebox{90}{SynthLoc} &
\rotatebox{90}{GoToSeq} &
\rotatebox{90}{SynthSeq} &
\rotatebox{90}{BossLevel} \\
\hline

Navigate a 6×6 room & \checkmark & \checkmark & \checkmark & \checkmark & \checkmark & \checkmark & \checkmark & \checkmark & \checkmark & \checkmark & \checkmark & \checkmark & \checkmark & \checkmark & \checkmark & \checkmark \\

Ignore grey box distractors & & \checkmark & \checkmark & \checkmark & \checkmark & \checkmark & & \checkmark & \checkmark & \checkmark & \checkmark & \checkmark & \checkmark & \checkmark & \checkmark & \checkmark \\

Ignore all distractors & & & \checkmark & \checkmark & \checkmark & \checkmark & & \checkmark & \checkmark & \checkmark & \checkmark & \checkmark & \checkmark & \checkmark & \checkmark & \checkmark \\

Navigate a 3×3 maze of 6×6 rooms & & & & & & & \checkmark & \checkmark & \checkmark & \checkmark & \checkmark & \checkmark & \checkmark & \checkmark & \checkmark & \checkmark \\

Move objects out of the way to navigate & & & & & & & & & & \checkmark & & \checkmark & \checkmark & & \checkmark & \checkmark \\

Unlock doors when explicitly instructed & & & & & & & & & & & & \checkmark & \checkmark & & \checkmark & \checkmark \\

Unlock doors when not explicitly stated & & & & & & & & & & & & & & & & \checkmark \\

Understand ``go to'' instructions & & & & \checkmark & & & & \checkmark & & & & \checkmark & \checkmark & \checkmark & \checkmark & \checkmark \\

Understand ``open'' instructions & & & & & & & & & & & \checkmark & \checkmark & \checkmark & & \checkmark & \checkmark \\

Understand ``pick up'' instructions & & & & & & \checkmark & & & \checkmark & \checkmark & & \checkmark & \checkmark & & \checkmark & \checkmark \\

Understand ``put'' instructions & & & & & \checkmark & & & & & & & \checkmark & \checkmark & & \checkmark & \checkmark \\

Understand relative location & & & & & & \checkmark & & & & & & & \checkmark & & \checkmark & \checkmark \\

Understand sequences of instructions & & & & & & & & & & & & & & \checkmark & \checkmark & \checkmark \\
\hline
\end{tabular}}
\end{table}

\subsubsection{Choice of Environment Formatter}
\label{apx:formatters}

The environment description can be serialised in more than one way, and prompt format is known to move reasoning performance on its own \citep{puerto-etal-2024-code, he2024doespromptformattingimpact}. We compared three formatters: a \emph{narrative} one that describes the grid in running English sentences, a \emph{structured} one that pairs a short prose header with bulleted key--value fields, and a \emph{JSON} one that puts the same fields in a JSON object. The comparison was run on \textbf{Predict} in an earlier round of experiments, with a different model set and a fixed prompting condition, so the absolute numbers in Table~\ref{tab:formatters} are not comparable with the results in Section~\ref{sec:predict-results}; only the ordering matters. The structured format was best on both metrics in that round, and it is what we use for every experiment reported in this paper. Our released code can regenerate the datasets under any of the three.

\begin{table}[h]
\centering
\small
\caption{Formatter comparison on \textbf{Predict}, averaged over the models evaluated in an earlier round. Manhattan distance is between the predicted and true final position, so lower is better. These figures come from a different model set under a fixed prompting condition and are not comparable with the results in Section~\ref{sec:predict-results}; only the ordering of the three formatters is used.}
\label{tab:formatters}
\begin{tabular}{lcc}
\toprule
Formatter & Success rate & Manhattan distance ($\downarrow$) \\
\midrule
Narrative        & 56.8\% & 2.34 \\
JSON             & 59.3\% & 2.65 \\
Structured text  & \textbf{61.1}\% & \textbf{2.24} \\
\bottomrule
\end{tabular}
\end{table}

\subsubsection{Sampling Uncertainty}
\label{apx:uncertainty}

Every rate we report is an average over cells, not over samples: pass@1 is the mean of the three samples within a cell, averaged over the cells in the relevant group, so the cell count is the effective sample size. Table~\ref{tab:cis} gives the half-width of a 95\% interval at representative group sizes spanning those that appear in this paper, computed at the worst case $p=0.5$. Near 0\% or 100\% the true interval is considerably narrower and asymmetric, and we recommend a Wilson interval there rather than the symmetric approximation.

\begin{table}[h]
\centering
\small
\caption{Worst-case half-width of a 95\% interval (percentage points) as a function of the number of cells in a group.}
\label{tab:cis}
\begin{tabular}{lrrrrrrrrr}
\toprule
Cells in group & 320 & 200 & 120 & 100 & 60 & 40 & 30 & 20 & 12 \\
\midrule
Half-width (pp) & 5.5 & 6.9 & 8.9 & 9.8 & 12.7 & 15.5 & 17.9 & 21.9 & 28.3 \\
\bottomrule
\end{tabular}
\end{table}

Two consequences follow for reading the tables. Differences of a few points between adjacent models in the global rankings are not resolvable, and the smallest buckets we report, such as the 10+ subgoal band of Table~\ref{tab:decompose-by-count} or the 3+ clause band of Table~\ref{tab:decompose-clauses}, carry intervals wide enough that only large effects should be read from them. The results we build arguments on, the horizon curves and the comprehension--precision gap, are separations well outside these intervals; on Decompose the smallest, Kimi-K2.5's 12.8-point comprehension--precision gap over all 320 cells, is more than twice the 5.5-point half-width.

\subsubsection{Prompt Structure and Querying Protocol}
\label{apx:prompts}

All main results were obtained with a zero-shot structured prompt. Every query is a single-turn prompt built from three parts.
\begin{enumerate}
  \item \textbf{Environment description}, produced by a structured formatter: a fixed natural-language header stating the world rules (room layout; the six available actions, namely left, right, forward, pickup, drop and toggle; the fact that only forward changes the agent's position; that occupied cells are impassable; and the $(x, y)$ coordinate convention with origin at the top-left corner), followed by a bulleted specification of the concrete instance: number and size of rooms, total grid size, the agent's initial position and facing direction, an exhaustive list of objects with type, color, position (and lock status for doors), and the mission string.
  \item \textbf{Task instruction}, specific to the benchmark task: for \textbf{Predict}, the action sequence to simulate and a request for the resulting agent state; for \textbf{Plan}, the goal of reaching the red ball and a request for an action sequence; for \textbf{Decompose}, the definitions of the five admissible subgoals (\texttt{OpenSubgoal}, \texttt{CloseSubgoal}, \texttt{PickupSubgoal}, \texttt{DropSubgoal}, and \texttt{GoNextToSubgoal (x, y)}) and a request for a chronologically ordered subgoal sequence.
  \item \textbf{Format anchor}, enforcing machine-parsable output: the model must end its response with a fixed phrase followed by the answer in a rigid syntax, namely \texttt{The agent's final state is: ((x, y), dir)} for \textbf{Predict}, \texttt{The LLM's action sequence is: action1, action2, ...} for \textbf{Plan}, and a \texttt{<START>}\ldots\texttt{<END>}-delimited subgoal list for \textbf{Decompose}. A deterministic parser extracts the answer from the last occurrence of the anchor.
\end{enumerate}
No in-context examples, chain-of-thought scaffolding, or system prompt were used, though models were free to reason before the anchor phrase.

All models were queried at temperature $t=0.7$ with $3$ independent samples per (model, level, seed) cell. Reasoning models were used at their default setting. For Kimi-K2.5, the default settings seems to reason for 4-6 minutes. All runs use the first 20 primes (2–71) as seeds for every level, with moonshotai/Kimi-K2.5, Qwen/Qwen2.5-72B-Instruct, meta-llama/Llama-4-Maverick-17B-128E-Instruct-FP8 and meta-llama/Llama-4-Scout-17B-16E-Instruct queried via DeepInfra and claude-opus-4-6, claude-sonnet-4-6 and gpt-5.4 via the Anthropic and OpenAI APIs directly, with a per-level output cap of 2048, 4096 or 8192 tokens (8192 throughout for Plan; 16384 for Kimi-K2.5).Each (model, task) block comprises 16 levels or level configurations $\times$ 20 seeds. We report pass@1 (mean over the 3 samples) and pass@3 (best of 3). Task-specific breakdowns are reported at pass@1 throughout, since several of their bins are too small to support a stable best-of-three estimate. For models exposing an extended-reasoning control (claude-opus-4-6, claude-sonnet-4-6, gpt-5.4, Kimi-K2.5), extended reasoning was left at its default setting in all runs. Responses from which the deterministic parser could not extract an answer are scored as failures. For continuous metrics, pass@3 selects the best of the three samples (maximum efficiency, minimum \texttt{ball\_distance}). Efficiency is reported conditional on success throughout: we pool it over the successful samples, so that completion rate and plan quality remain separate quantities. \texttt{ball\_distance} is averaged per cell and then over cells, and is defined on the responses that yield a final agent position.  Response length and latency are computed over the responses returned; on the Ultra track this is 212 of 240 samples for Kimi-K2.5. Its 28 unreturned samples count as failures in every success rate and are excluded from Table~\ref{tab:plan-blocks}.

\begin{tcolorbox}[colback=verylightgray, title={Example \textbf{Plan} prompt}]
\ttfamily\footnotesize
An agent is in a grid world consisting of one or more rooms. [...] The agent can perform 6 actions: left (turn left), right (turn right), forward (move forward), pickup (pickup an object), drop (drop an object), and toggle (open/close a door). Only the forward action changes the agent's position [...] Using a coordinate system where the (0, 0) position is the top-left corner of the grid world [...]

\medskip
These are the specifics regarding this environment:

\medskip
Number of rooms: 1\\
Size of each room (including walls): 8x8\\
Total grid size: 8x8\\
Agent initial position: (5, 3)\\
Agent facing direction: south (toward (5, 4))\\
Objects in environment:\\
ball, color=red, position=(5, 1)\\
ball, color=grey, position=(3, 3)\\
Mission: 'go to the red ball'

\medskip
So, your goal is to reach the red ball, and you need to determine the sequence of actions to get there. At the end of your response, write the action sequence in the following way action1, action2, action3 [...] Begin this section with the phrase: 'The LLM's action sequence is: '.
\end{tcolorbox}
\subsubsection{Custom Environment for the Plan Task}
\label{app:custom_env_plan_task}
\begin{figure}
    \centering
    \includegraphics[width=0.5\linewidth]{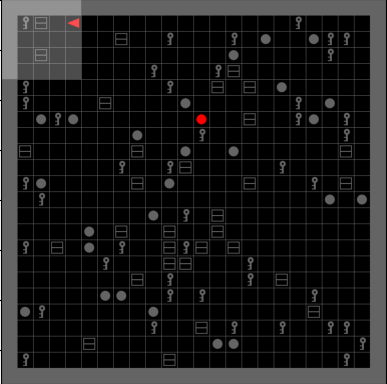}
    \caption{An instance of the custom Plan environment on a Large ($24\times24$) grid.}
    \label{fig:customGoToRedBallGrey}
\end{figure}

Since the Plan task exclusively involves \texttt{GoNextTo} subgoals, we designed a custom environment derived from the BabyAI level \texttt{GoToRedBallGrey}. Both the grid size and the number of distractor objects are configurable. Distractors are items placed so as to obstruct the agent's path, forcing it to navigate around them. We use four sizes: Small ($8\times8$), Medium ($16\times16$), Large ($24\times24$) and Ultra ($32\times32$). Figure~\ref{fig:customGoToRedBallGrey} shows an example of such a grid.

We name these environments \texttt{CustomBabyAI-GoToRedBall-\{size\}-\{number of distractors\}Dists-v0} depending on the size and the number of distractors.

\subsubsection{Why OmniBot and LLM Decompositions Are Not Compared Directly}

One could evaluate the LLM's decomposition by comparing its proposed subgoal stack with that of the OmniBot. This comparison is not fair, because the two systems produce and revise their subgoals in different ways.

The OmniBot interacts with the environment in real time and can dynamically update its subgoal stack as it gathers new information or encounters obstacles. In contrast, the LLM generates a fixed list of subgoals in advance, without the ability to revise them during execution. Since the LLM has full visibility of the grid at generation time, it is theoretically possible to anticipate all necessary steps and contingencies upfront.

Even assuming a perfect LLM with ideal reasoning capabilities, its generated subgoals will generally not match those of the OmniBot. The mismatch is caused by the LLM having to commit to a plan in advance while the OmniBot adapts as it goes, so it does not reflect the quality of the LLM's reasoning.

\begin{figure}[h]
  \centering
  \includegraphics[width=0.5\textwidth]{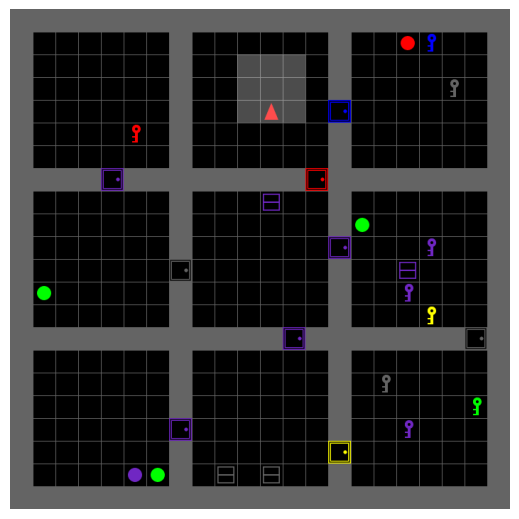}
  \caption{BabyAI GoTo level with seed 54. The task is to ``go to the blue key''. This example illustrates why the OmniBot and a perfect LLM may produce different subgoal sequences.}
  \label{fig:goto54}
\end{figure}

Consider the example shown in Figure~\ref{fig:goto54}, where the agent must go to the blue key. The OmniBot begins the episode with the subgoal stack:
\[
\texttt{[GoNextToSubgoal(blue key)]}
\]
As the episode progresses, the OmniBot moves toward the blue key. Upon reaching the blue door blocking the path, it appends a new subgoal:
\[
\texttt{[GoNextToSubgoal(blue key), OpenSubgoal]}
\]
After opening the door, the OmniBot removes the now-completed \texttt{OpenSubgoal}, returning the stack to its original form before completing the task.

In contrast, the LLM must anticipate the need to open the door and generate the complete subgoal sequence upfront:
\[
\texttt{[GoNextToSubgoal(blue key), OpenSubgoal, GoNextToSubgoal(blue door)]}
\]
The difference arises because the OmniBot follows a dynamic subgoal policy, while the LLM must commit to a static plan.

Therefore, a direct comparison between the LLM's subgoals and those of the OmniBot is misleading for two reasons:
\begin{enumerate}
    \item The OmniBot's subgoal stack is dynamic and evolves over time.
    \item A perfect LLM will include subgoals that are never explicitly added by the OmniBot, since they are only needed transiently and are handled reactively by the bot.
\end{enumerate}

Directly matching subgoal stacks would therefore penalise even optimal LLM predictions, so such a metric is unsuitable for evaluating decomposition quality.

\subsubsection{Why the OmniBot Always Appends a Drop}
\label{apx:drop}

The OmniBot revises its own subgoal stack while an episode runs. CR allows a model to rely on these revisions, and PR does not. A typical UnblockPickup episode starts from the default expansion of the mission,
\[
\texttt{[DropSubgoal, PickupSubgoal, GoNextToSubgoal(green ball)]},
\]
and then grows and shrinks as the bot meets obstacles it has to clear. Partway through it is holding
\begin{align*}
&\texttt{[DropSubgoal, PickupSubgoal, GoNextToSubgoal(green ball), }\\
&\texttt{DropSubgoal, GoNextToSubgoal(20,12), PickupSubgoal]},
\end{align*}
having decided to move a blocking object out of the doorway, and later it inserts an \texttt{OpenSubgoal} when it reaches a closed door. Each inserted subgoal is popped once satisfied, so the stack returns to its original shape before the mission completes. A model writing a static plan has to anticipate all of this in advance.

One element of the default expansion needs explanation, because it looks redundant and a model would have no reason to guess it. Whenever a mission requires picking an object up, the bot places a \texttt{DropSubgoal} at the bottom of the stack, so the agent puts down whatever it is carrying once the pickup has served its purpose. In the episode above the mission is only to pick up the green ball, and the trailing drop appears to accomplish nothing.

\begin{figure}[h]
  \centering
  \includegraphics[width=0.45\textwidth]{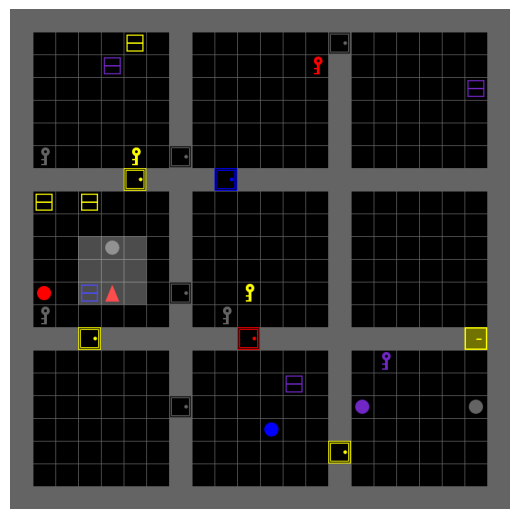}
  \caption{SynthSeq, seed 166. The mission is ``pick up a grey ball and go to the ball in front of you, then go to a box and put a purple box next to the red door''. Nothing in the mission asks the agent to put the grey ball down, but an agent that keeps holding it makes the second clause impossible to satisfy.}
  \label{fig:synthseq166}
\end{figure}

It is a safety mechanism, and Figure~\ref{fig:synthseq166} shows why it is needed. The mission there is to pick up a grey ball and then go to the ball in front of the agent, before two further clauses. If the agent keeps holding the grey ball it picked up, the ball is no longer on the grid in front of it, and the second clause becomes unsatisfiable through no fault of the plan that produced it. Dropping the object as soon as it is no longer required keeps the world in a state where later clauses can still be satisfied. The consequence for our metrics is that PR requires a model to reproduce a convention of the executor rather than a property of the task, which is part of why we read a PR of zero as the absence of a self-sufficient decomposition only; the decomposition may still be usable.

\subsubsection{The Evaluation Process}
\label{app:eval_process_decompose}
As direct comparison between subgoals generated by the LLM and those used by the OmniBot is not meaningful (as discussed above), we propose a new evaluation strategy tailored specifically to the \benchmark Decompose task. This procedure involves the following steps:

\begin{enumerate}
\item We prompt the LLM to decompose the level's instruction into a sequence of subgoals.
\item The bot is initialised with this sequence of subgoals.
\item The bot then executes an episode. During execution, we record the number of additional subgoals the bot needs to append in order to complete the task (whether by success, reaching the maximum number of steps, or depleting the subgoal stack).
\item For each episode, we record the success status and the number of subgoals added. These values are used to compute three metrics: Comprehension Rate (CR), Precision Rate (PR), and Assistance Curve Integral (ACI).
\end{enumerate}
\subsubsection{What Comprehension Rate Measures}

The Comprehension Rate (CR) measures the proportion of episodes successfully completed, regardless of how many subgoals the OmniBot needs to add during execution. To achieve a high CR, it suffices for the LLM to generate a plausible decomposition of the instruction into subgoals. As noted in Section~\ref{sec:omnibot}, such a decomposition usually enables the OmniBot to solve the level successfully, even if the initial stack does not account for the specific obstacles present in the environment. For example, in Figure~\ref{fig:goto54}, the LLM only needs to understand that the level's mission \texttt{go to the blue key} can be reformulated as the subgoal \texttt{GoNextTo(blue key)} to positively contribute to the CR metric.

This makes CR a relatively undemanding metric: it allows the OmniBot to freely intervene and compensate for incomplete or imprecise LLM outputs, and it does not require the LLM to reason about the environment's layout. Rather, the LLM only needs to reformulate the instruction mechanically into a sequence of subgoals from a predefined set. For this reason, we interpret CR as a measure of the LLM's task comprehension: its ability to understand and translate a high-level instruction into a structured goal representation.

\subsubsection{What Precision Rate Measures}

The \textit{Precision Rate} (PR) quantifies the proportion of episodes successfully completed without the OmniBot needing to insert any additional subgoals during execution. A high PR indicates that the language model has fully anticipated all necessary steps and obstacles, explicitly encoding them as subgoals in its initial output.

For instance, in Figure~\ref{fig:goto54}, the shortest correct decomposition is \texttt{[GoNextToSubgoal(blue key), OpenSubgoal, GoNextToSubgoal(blue door)]}, which the model must generate unaided to count toward PR.

\begin{figure}[h]
  \centering
  \includegraphics[width=0.5\textwidth]{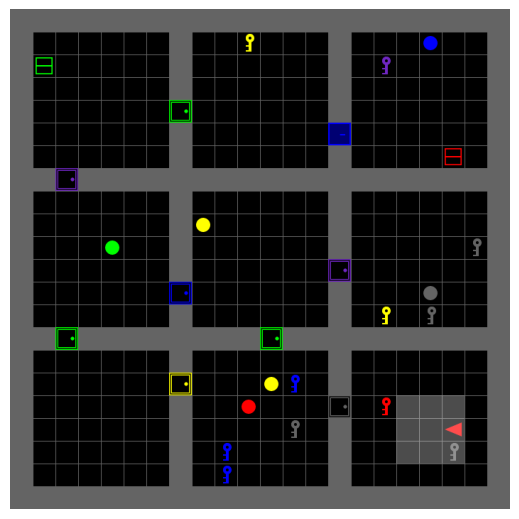}
  \caption{BabyAI BossLevel (seed 81). The task is to ``open a green door, then put a yellow ball next to the red box''. This mission illustrates a clear contrast in planning complexity: with unlimited assistance from the OmniBot, a concise list of just 6 subgoals is sufficient for success, whereas achieving the same outcome without any assistance requires an extended sequence of 22 subgoals.}
  \label{fig:bosslevel81}
\end{figure}

PR is especially demanding in complex scenarios that involve multiple intermediate steps. Consider the example in Figure~\ref{fig:bosslevel81}: with OmniBot assistance, the following short sequence is enough to complete the mission:
\begin{center}
\texttt{[DropSubgoal, GoNextToSubgoal(red box), PickupSubgoal, GoNextToSubgoal(yellow ball), OpenSubgoal, GoNextToSubgoal(green door)]}.
\end{center}
However, completing the task without any subgoal insertion by the bot requires the model to generate a significantly longer and more detailed sequence:
\begin{center}
\texttt{[DropSubgoal, GoNextToSubgoal(red box), PickupSubgoal, GoNextToSubgoal(yellow ball), OpenSubgoal, GoNextToSubgoal(blue door), PickupSubgoal, GoNextToSubgoal(blue key), DropSubgoal, GoNextToSubgoal(drop position), OpenSubgoal, GoNextToSubgoal(green door), OpenSubgoal, GoNextToSubgoal(purple door), OpenSubgoal, GoNextToSubgoal(blue door), PickupSubgoal, GoNextToSubgoal(yellow ball), OpenSubgoal, GoNextToSubgoal(green door), OpenSubgoal, GoNextToSubgoal(grey door)]}.
\end{center}

\subsubsection{What the Assistance Curve Integral Measures}

While the \textit{Comprehension Rate} measures the success rate with unlimited help from the OmniBot, and the \textit{Precision Rate} measures success without any assistance, the \textit{Assistance Curve Integral} (ACI) provides a smooth interpolation between these two extremes. Rather than selecting an arbitrary threshold $k$, ACI integrates the success rate as a function of the number of subgoals added, from $0$ to a predefined \textit{limit}. 
This limit is $M_{\text{help}}$, the size of the OmniBot's own from-scratch decomposition of the mission, which depends only on the level and the seed (Section~\ref{sec:omnibot}).
The integral is then normalised by dividing by \textit{limit}, such that a perfect model achieves an ACI of 1.

A high ACI indicates that the model typically requires little assistance to solve tasks, whereas a low ACI suggests heavy reliance on OmniBot interventions.

\begin{figure}[h]
  \centering
  \includegraphics[width=0.5\textwidth]{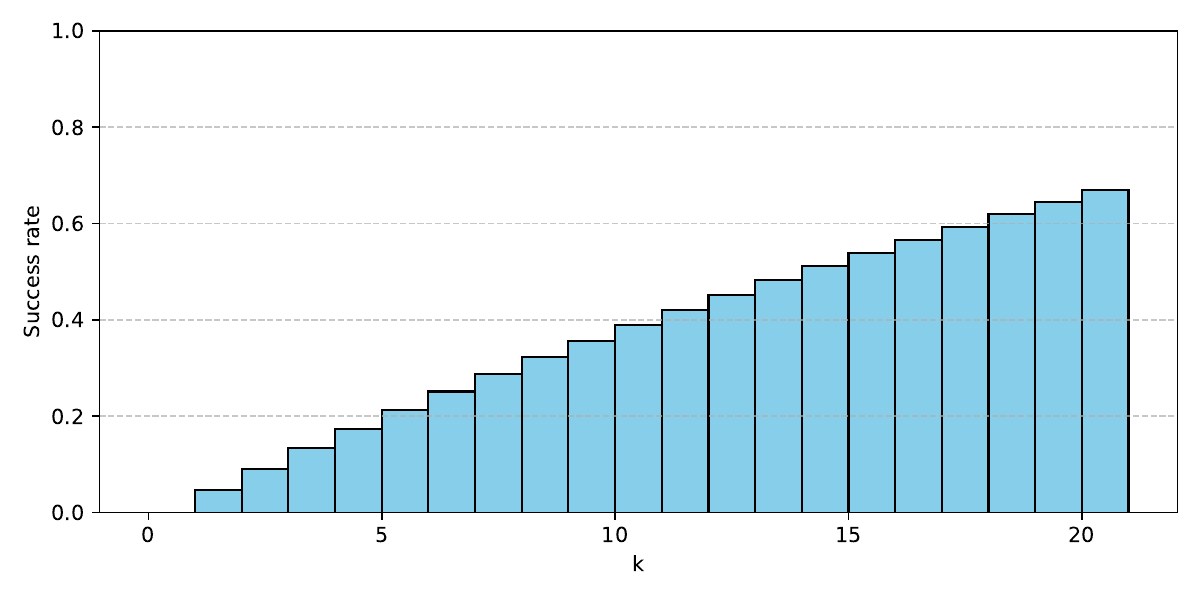}
  \caption{Example bar chart of the success rate as a function of the number of subgoals added by the OmniBot. The blue area represents the unnormalised ACI.}
  \label{fig:aci_curve}
\end{figure}
\subsection{Plan Failure by Grid Size, Obstacle Density and Solution Length}
\label{apx:plan-analysis}

This section supports the analysis of Section~\ref{sec:plan} with the full per-model statistics behind Table~\ref{tab:plan-by-length}. The Plan track varies two geometric factors. The first is grid size (Small $8\times8$, Medium $16\times16$, Large $24\times24$, Ultra $32\times32$; effective interior 36, 196, 484 and 900 cells), the second the number of distractor objects (Small 4--7, Medium 20--60, Large 60--120, Ultra 120--180). Together they yield 16 level configurations with obstacle densities between 10\% and 31\% of interior cells. Because larger grids are evaluated with more distractors, size and density are confounded across levels. The analyses below separate them and show that both act through a third variable, the length of the optimal solution (mean oracle plan length: 6.6 actions on Small, 13.6 on Medium, 18.6 on Large, 24.4 on Ultra).

\begin{table}[h]
\centering
\small
\caption{Plan by grid size: pass@1 Success Rate (Efficiency).}
\label{tab:plan-by-size}
\begin{tabular}{l rrrr}
\toprule
Model & Small ($8\times8$) & Medium ($16\times16$) & Large ($24\times24$) & Ultra ($32\times32$) \\
\midrule
claude-opus-4-6   & 99.6\% (0.96) & 78.8\% (0.97) & 69.6\% (0.96) & 60.8\% (0.93) \\
Kimi-K2.5         & 89.2\% (0.89) & 77.9\% (0.90) & 70.8\% (0.91) & 61.7\% (0.93) \\
claude-sonnet-4-6 & 97.1\% (0.96) & 71.7\% (0.96) & 64.2\% (0.93) & 50.8\% (0.93) \\
gpt-5.4           & 51.7\% (0.91) & 27.5\% (0.83) & 24.6\% (0.87) & 22.5\% (0.88) \\
Llama-4-Maverick  & 35.4\% (0.77) & 16.7\% (0.79) &  9.2\% (0.79) &  5.8\% (0.82) \\
Llama-4-Scout     & 17.9\% (0.70) &  5.8\% (0.72) &  2.9\% (0.79) &  0.8\% (0.52) \\
Qwen2.5-72B       & 15.0\% (0.44) &  2.5\% (0.13) &  2.5\% (0.69) &  0.4\% (0.80) \\
\bottomrule
\end{tabular}
\end{table}

\textbf{Success stratified by oracle path length within each grid size.} Table~\ref{tab:plan-terciles} splits each grid size's instances into terciles of oracle plan length. Within every non-Small size, and for the weaker models within Small as well, path length stratifies success sharply while grid size is held fixed, so the full collapse of Table~\ref{tab:plan-by-size} is reproduced at constant grid size. claude-opus-4-6 is saturated on Small (100/99/100), which is where all its paths lie below its horizon.

\begin{table}[h]
\centering
\footnotesize
\caption{Plan: pass@1 success rate (\%) by within-size tercile of oracle plan length (short / mid / long).}
\label{tab:plan-terciles}
\begin{tabular}{l cccc}
\toprule
Model & Small & Medium & Large & Ultra \\
\midrule
claude-opus-4-6   & 100 / 99 / 100 & 100 / 86 / 41 & 99 / 77 / 28 & 88 / 64 / 25 \\
claude-sonnet-4-6 & 100 / 99 / 83  & 97 / 72 / 38  & 95 / 64 / 29 & 81 / 52 / 14 \\
Kimi-K2.5         & 93 / 87 / 89   & 91 / 83 / 55  & 85 / 80 / 43 & 85 / 56 / 42 \\
gpt-5.4           & 80 / 44 / 14   & 57 / 13 / 8   & 52 / 16 / 4  & 46 / 12 / 7  \\
Llama-4-Maverick  & 54 / 33 / 0    & 37 / 7 / 3    & 27 / 0 / 0   & 14 / 0 / 3   \\
Llama-4-Scout     & 36 / 11 / 0    & 16 / 0 / 0    & 9 / 0 / 0    & 2 / 0 / 0    \\
Qwen2.5-72B       & 32 / 8 / 0     & 6 / 1 / 0     & 7 / 0 / 0    & 1 / 0 / 0    \\
\bottomrule
\end{tabular}
\end{table}

\textbf{Density and size correlations.} Table~\ref{tab:plan-correlations} reports cell-level Pearson correlations between obstacle density and completion within each grid size, and level-level correlations of completion with density and with log grid size across the 16 configurations. First, within-size density effects appear only where a model operates on the sloped region of its horizon curve: they are clear for the strong models on Medium--Ultra ($r=-0.21$ to $-0.41$), absent or weak on Small (claude-opus-4-6 $+0.05$, Kimi-K2.5 $-0.02$; claude-sonnet-4-6 shows a mild $-0.23$), and are weak and inconsistent in sign for the floor models (Llama-4-Scout, Qwen2.5-72B: $-0.24$ to $+0.14$). Second, the level-level correlation with log grid size broadly strengthens as models weaken ($-0.71$ for Kimi-K2.5 to $-0.89$ for Qwen2.5-72B): once a model's horizon is short, the baseline path length imposed by grid size dominates the other factors. Third, sample-level point-biserial correlations between success and oracle length are $-0.63$ (claude-opus-4-6), $-0.62$ (claude-sonnet-4-6), $-0.46$ (gpt-5.4), $-0.41$ (Llama-4-Maverick), $-0.40$ (Kimi-K2.5), $-0.29$ (Llama-4-Scout) and $-0.24$ (Qwen2.5-72B); the coefficient shrinks at the extremes because a flatter profile (Kimi-K2.5) and floors (Llama-4-Scout, Qwen2.5-72B) compress the observable variation.

\begin{table}[h]
\centering
\footnotesize
\caption{Correlations between completion (pass@1) and the geometric factors. Left: cell-level density--SR correlation within each grid size. Right: level-level correlation across the 16 configurations.}
\label{tab:plan-correlations}
\begin{tabular}{l cccc cc}
\toprule
& \multicolumn{4}{c}{Within-size density--SR $r$} & \multicolumn{2}{c}{Across 16 levels} \\
\cmidrule(lr){2-5}\cmidrule(lr){6-7}
Model & Small & Medium & Large & Ultra & Density & log grid size \\
\midrule
claude-opus-4-6   & $+0.05$ & $-0.35$ & $-0.33$ & $-0.29$ & $-0.56$ & $-0.79$ \\
claude-sonnet-4-6 & $-0.23$ & $-0.34$ & $-0.40$ & $-0.21$ & $-0.55$ & $-0.82$ \\
Kimi-K2.5         & $-0.02$ & $-0.41$ & $-0.34$ & $-0.24$ & $-0.56$ & $-0.71$ \\
gpt-5.4           & $-0.13$ & $-0.18$ & $-0.11$ & $-0.02$ & $-0.49$ & $-0.80$ \\
Llama-4-Maverick  & $-0.11$ & $-0.22$ & $-0.15$ & $+0.08$ & $-0.42$ & $-0.86$ \\
Llama-4-Scout     & $-0.12$ & $+0.06$ & $-0.24$ & $+0.14$ & $-0.27$ & $-0.87$ \\
Qwen2.5-72B       & $+0.08$ & $+0.10$ & $-0.00$ & $-0.05$ & $-0.12$ & $-0.89$ \\
\bottomrule
\end{tabular}
\end{table}

\textbf{Matched obstacle counts.} Holding the number of objects fixed while varying grid size dissociates prompt length from geometry. For claude-opus-4-6, Large-120 yields 48.3\% against 80.0\% on Ultra-120, and for Kimi-K2.5, Medium-60 yields 61.7\% against 83.3\% on Large-60. The object list, and with it the prompt, is almost unchanged in both comparisons, yet the larger and less dense grid is much easier. For Llama-4-Maverick the same comparisons are flat or slightly reversed (Medium-60: 16.7\% against Large-60: 13.3\%), as horizon mediation predicts: Large-grid paths exceed Llama-4-Maverick's horizon whatever the density, so the density relief that helps the strong models does nothing for it. Across models, then, neither geometric factor accounts for the matched-count comparisons, while solution length does.

\textbf{Sample consistency.} Table~\ref{tab:plan-consistency} gives the distribution of per-cell outcomes over the three samples. The profiles separate a deterministic component of failure (0/3 cells: instances beyond the model's horizon) from a variance component (1/3 and 2/3 cells), and the balance between them depends on the model but not on its overall strength: the Claude models are close to bimodal, gpt-5.4 and Kimi-K2.5 carry a large stochastic mass, and the weak models fail almost everywhere deterministically, with occasional successes.

\begin{table}[h]
\centering
\footnotesize
\caption{Plan: distribution of successes per 3-sample cell (320 cells per model), the share of 0/3 cells on Ultra, and the resampling gain (pass@3 $-$ pass@1, percentage points).}
\label{tab:plan-consistency}
\begin{tabular}{l rrrr r r}
\toprule
Model & 3/3 & 2/3 & 1/3 & 0/3 & 0/3 (Ultra) & p@3 $-$ p@1 \\
\midrule
claude-opus-4-6   & 67.2\% & 11.2\% &  7.5\% & 14.1\% & 28\% &  8.7 \\
claude-sonnet-4-6 & 57.2\% & 13.8\% & 13.8\% & 15.3\% & 29\% & 13.8 \\
Kimi-K2.5         & 51.6\% & 26.6\% & 16.9\% &  5.0\% &  8\% & 20.1 \\
gpt-5.4           & 17.5\% & 11.6\% & 19.1\% & 51.9\% & 59\% & 16.5 \\
Llama-4-Maverick  &  7.8\% &  9.4\% &  8.1\% & 74.7\% & 91\% &  8.5 \\
Llama-4-Scout     &  0.6\% &  4.7\% &  9.4\% & 85.3\% & 98\% &  7.8 \\
Qwen2.5-72B       &  1.6\% &  2.8\% &  5.0\% & 90.6\% & 99\% &  4.3 \\
\bottomrule
\end{tabular}
\end{table}

\textbf{Per-level results.} Tables~\ref{tab:plan-levels-claudeopus46}--\ref{tab:plan-levels-qwen2572b} report, for each model, completion, efficiency conditional on success, and \texttt{ball\_distance} for each of the 16 level configurations ($n=60$ samples per level configuration). Efficiency conditional on success varies little per model across sizes and densities (claude-opus-4-6: 0.92--0.98; claude-sonnet-4-6: 0.87--0.97; Kimi-K2.5: 0.85--0.95; gpt-5.4: 0.81--0.95); cells with very few successes should not be read as measurements, since one sample can set the value. The two Qwen2.5-72B cells at 0.003 and 0.001 are a different artifact: the model emits a degenerate 4{,}092-action sequence that exhausts the output cap, the episode happens to reach the goal partway through it, and the ratio divides the oracle's three to eleven actions by the full emitted length. Three caveats apply throughout. At $n=60$ per cell, several small non-monotonicities (e.g.\ Llama-4-Maverick: Medium-50 at 5.0\% vs.\ Medium-60 at 16.7\%; Llama-4-Scout: Medium-20 at 3.3\% vs.\ Medium-40 at 10.0\%) are within sampling noise and should not be over-read; a 95\% binomial interval at $n=20$ cells spans roughly $\pm 22$ points at mid-range rates. The 46+ oracle-length bin of the raw data contains only $n=9$ samples and is merged into the 31+ bin of Table~\ref{tab:plan-by-length}. Finally, the \texttt{ball\_distance} columns average over all samples including successes (which contribute 0), so failure-conditional distances are larger than the reported means.

\begin{table}[h]
\centering\footnotesize
\caption{Plan per level configuration: claude-opus-4-6. SR = success rate; Eff$\mid$S = efficiency conditional on success; bd = \texttt{ball\_distance}. $n=60$ samples (20 seeds $\times$ 3) per row.}
\label{tab:plan-levels-claudeopus46}
\begin{tabular}{l r r r r}
\toprule
Level & Density & SR p@1 & Eff$\mid$S p@1 & bd p@1 \\
\midrule
Small-4 & 11.1\% & 100.0\% & 0.95 & 0.00 \\
Small-5 & 13.9\% & 98.3\% & 0.95 & 0.03 \\
Small-6 & 16.7\% & 100.0\% & 0.95 & 0.00 \\
Small-7 & 19.4\% & 100.0\% & 0.98 & 0.00 \\
Medium-20 & 10.2\% & 98.3\% & 0.96 & 0.07 \\
Medium-40 & 20.4\% & 81.7\% & 0.97 & 1.95 \\
Medium-50 & 25.5\% & 70.0\% & 0.98 & 1.97 \\
Medium-60 & 30.6\% & 65.0\% & 0.97 & 1.53 \\
Large-60 & 12.4\% & 85.0\% & 0.96 & 0.85 \\
Large-80 & 16.5\% & 71.7\% & 0.96 & 1.97 \\
Large-100 & 20.7\% & 73.3\% & 0.96 & 2.25 \\
Large-120 & 24.8\% & 48.3\% & 0.97 & 5.12 \\
Ultra-120 & 13.3\% & 80.0\% & 0.92 & 2.50 \\
Ultra-140 & 15.6\% & 63.3\% & 0.92 & 3.97 \\
Ultra-160 & 17.8\% & 55.0\% & 0.94 & 6.82 \\
Ultra-180 & 20.0\% & 45.0\% & 0.96 & 8.68 \\
\bottomrule
\end{tabular}
\end{table}

\begin{table}[h]
\centering\footnotesize
\caption{Plan per level configuration: claude-sonnet-4-6. SR = success rate; Eff$\mid$S = efficiency conditional on success; bd = \texttt{ball\_distance}. $n=60$ samples (20 seeds $\times$ 3) per row.}
\label{tab:plan-levels-claudesonnet46}
\begin{tabular}{l r r r r}
\toprule
Level & Density & SR p@1 & Eff$\mid$S p@1 & bd p@1 \\
\midrule
Small-4 & 11.1\% & 100.0\% & 0.95 & 0.00 \\
Small-5 & 13.9\% & 100.0\% & 0.96 & 0.00 \\
Small-6 & 16.7\% & 95.0\% & 0.95 & 0.13 \\
Small-7 & 19.4\% & 93.3\% & 0.97 & 0.15 \\
Medium-20 & 10.2\% & 91.7\% & 0.97 & 0.23 \\
Medium-40 & 20.4\% & 76.7\% & 0.96 & 1.72 \\
Medium-50 & 25.5\% & 56.7\% & 0.96 & 2.87 \\
Medium-60 & 30.6\% & 61.7\% & 0.97 & 1.92 \\
Large-60 & 12.4\% & 83.3\% & 0.93 & 1.65 \\
Large-80 & 16.5\% & 68.3\% & 0.94 & 3.47 \\
Large-100 & 20.7\% & 66.7\% & 0.93 & 2.45 \\
Large-120 & 24.8\% & 38.3\% & 0.87 & 6.68 \\
Ultra-120 & 13.3\% & 61.7\% & 0.93 & 4.02 \\
Ultra-140 & 15.6\% & 56.7\% & 0.92 & 5.72 \\
Ultra-160 & 17.8\% & 45.0\% & 0.95 & 7.70 \\
Ultra-180 & 20.0\% & 40.0\% & 0.93 & 8.85 \\
\bottomrule
\end{tabular}
\end{table}

\begin{table}[h]
\centering\footnotesize
\caption{Plan per level configuration: Kimi-K2.5. SR = success rate; Eff$\mid$S = efficiency conditional on success; bd = \texttt{ball\_distance}. $n=60$ samples (20 seeds $\times$ 3) per row.}
\label{tab:plan-levels-kimik25}
\begin{tabular}{l r r r r}
\toprule
Level & Density & SR p@1 & Eff$\mid$S p@1 & bd p@1 \\
\midrule
Small-4 & 11.1\% & 88.3\% & 0.92 & 0.03 \\
Small-5 & 13.9\% & 91.7\% & 0.88 & 0.12 \\
Small-6 & 16.7\% & 88.3\% & 0.88 & 0.27 \\
Small-7 & 19.4\% & 88.3\% & 0.90 & 0.07 \\
Medium-20 & 10.2\% & 91.7\% & 0.93 & 0.02 \\
Medium-40 & 20.4\% & 86.7\% & 0.88 & 1.15 \\
Medium-50 & 25.5\% & 71.7\% & 0.90 & 1.52 \\
Medium-60 & 30.6\% & 61.7\% & 0.85 & 2.13 \\
Large-60 & 12.4\% & 83.3\% & 0.93 & 0.33 \\
Large-80 & 16.5\% & 76.7\% & 0.93 & 2.88 \\
Large-100 & 20.7\% & 71.7\% & 0.91 & 2.32 \\
Large-120 & 24.8\% & 51.7\% & 0.87 & 6.82 \\
Ultra-120 & 13.3\% & 75.0\% & 0.92 & 2.73 \\
Ultra-140 & 15.6\% & 58.3\% & 0.94 & 5.12 \\
Ultra-160 & 17.8\% & 63.3\% & 0.95 & 5.33 \\
Ultra-180 & 20.0\% & 50.0\% & 0.91 & 7.23 \\
\bottomrule
\end{tabular}
\end{table}

\begin{table}[h]
\centering\footnotesize
\caption{Plan per level configuration: gpt-5.4. SR = success rate; Eff$\mid$S = efficiency conditional on success; bd = \texttt{ball\_distance}. $n=60$ samples (20 seeds $\times$ 3) per row.}
\label{tab:plan-levels-gpt54}
\begin{tabular}{l r r r r}
\toprule
Level & Density & SR p@1 & Eff$\mid$S p@1 & bd p@1 \\
\midrule
Small-4 & 11.1\% & 61.7\% & 0.87 & 0.72 \\
Small-5 & 13.9\% & 56.7\% & 0.92 & 0.83 \\
Small-6 & 16.7\% & 36.7\% & 0.89 & 1.33 \\
Small-7 & 19.4\% & 51.7\% & 0.95 & 0.97 \\
Medium-20 & 10.2\% & 41.7\% & 0.81 & 3.13 \\
Medium-40 & 20.4\% & 25.0\% & 0.81 & 4.60 \\
Medium-50 & 25.5\% & 15.0\% & 0.81 & 4.60 \\
Medium-60 & 30.6\% & 28.3\% & 0.88 & 4.07 \\
Large-60 & 12.4\% & 26.7\% & 0.84 & 5.33 \\
Large-80 & 16.5\% & 28.3\% & 0.84 & 5.25 \\
Large-100 & 20.7\% & 28.3\% & 0.91 & 5.85 \\
Large-120 & 24.8\% & 15.0\% & 0.91 & 7.67 \\
Ultra-120 & 13.3\% & 25.0\% & 0.85 & 7.95 \\
Ultra-140 & 15.6\% & 16.7\% & 0.89 & 10.20 \\
Ultra-160 & 17.8\% & 30.0\% & 0.91 & 10.02 \\
Ultra-180 & 20.0\% & 18.3\% & 0.86 & 11.65 \\
\bottomrule
\end{tabular}
\end{table}

\begin{table}[h]
\centering\footnotesize
\caption{Plan per level configuration: Llama-4-Maverick. SR = success rate; Eff$\mid$S = efficiency conditional on success; bd = \texttt{ball\_distance}. $n=60$ samples (20 seeds $\times$ 3) per row.}
\label{tab:plan-levels-llama4maverick}
\begin{tabular}{l r r r r}
\toprule
Level & Density & SR p@1 & Eff$\mid$S p@1 & bd p@1 \\
\midrule
Small-4 & 11.1\% & 35.0\% & 0.78 & 2.17 \\
Small-5 & 13.9\% & 45.0\% & 0.78 & 1.70 \\
Small-6 & 16.7\% & 36.7\% & 0.75 & 2.27 \\
Small-7 & 19.4\% & 25.0\% & 0.75 & 2.17 \\
Medium-20 & 10.2\% & 31.7\% & 0.82 & 4.43 \\
Medium-40 & 20.4\% & 13.3\% & 0.81 & 6.32 \\
Medium-50 & 25.5\% & 5.0\% & 0.62 & 6.45 \\
Medium-60 & 30.6\% & 16.7\% & 0.78 & 5.43 \\
Large-60 & 12.4\% & 13.3\% & 0.80 & 8.93 \\
Large-80 & 16.5\% & 10.0\% & 0.74 & 9.50 \\
Large-100 & 20.7\% & 11.7\% & 0.81 & 8.73 \\
Large-120 & 24.8\% & 1.7\% & 0.83 & 11.88 \\
Ultra-120 & 13.3\% & 3.3\% & 1.00 & 12.48 \\
Ultra-140 & 15.6\% & 3.3\% & 0.80 & 15.47 \\
Ultra-160 & 17.8\% & 11.7\% & 0.86 & 12.42 \\
Ultra-180 & 20.0\% & 5.0\% & 0.60 & 15.25 \\
\bottomrule
\end{tabular}
\end{table}

\begin{table}[h]
\centering\footnotesize
\caption{Plan per level configuration: Llama-4-Scout. SR = success rate; Eff$\mid$S = efficiency conditional on success; bd = \texttt{ball\_distance}. $n=60$ samples (20 seeds $\times$ 3) per row.}
\label{tab:plan-levels-llama4scout}
\begin{tabular}{l r r r r}
\toprule
Level & Density & SR p@1 & Eff$\mid$S p@1 & bd p@1 \\
\midrule
Small-4 & 11.1\% & 18.3\% & 0.70 & 2.97 \\
Small-5 & 13.9\% & 25.0\% & 0.71 & 2.65 \\
Small-6 & 16.7\% & 16.7\% & 0.72 & 2.90 \\
Small-7 & 19.4\% & 11.7\% & 0.65 & 3.15 \\
Medium-20 & 10.2\% & 3.3\% & 0.65 & 6.82 \\
Medium-40 & 20.4\% & 10.0\% & 0.74 & 8.72 \\
Medium-50 & 25.5\% & 1.7\% & 0.75 & 8.30 \\
Medium-60 & 30.6\% & 8.3\% & 0.71 & 7.05 \\
Large-60 & 12.4\% & 8.3\% & 0.83 & 12.47 \\
Large-80 & 16.5\% & 3.3\% & 0.68 & 13.55 \\
Large-100 & 20.7\% & 0.0\% & --- & 11.68 \\
Large-120 & 24.8\% & 0.0\% & --- & 12.58 \\
Ultra-120 & 13.3\% & 0.0\% & --- & 18.03 \\
Ultra-140 & 15.6\% & 0.0\% & --- & 19.57 \\
Ultra-160 & 17.8\% & 1.7\% & 0.44 & 17.55 \\
Ultra-180 & 20.0\% & 1.7\% & 0.60 & 18.83 \\
\bottomrule
\end{tabular}
\end{table}

\begin{table}[h]
\centering\footnotesize
\caption{Plan per level configuration: Qwen2.5-72B. SR = success rate; Eff$\mid$S = efficiency conditional on success; bd = \texttt{ball\_distance}. $n=60$ samples (20 seeds $\times$ 3) per row.}
\label{tab:plan-levels-qwen2572b}
\begin{tabular}{l r r r r}
\toprule
Level & Density & SR p@1 & Eff$\mid$S p@1 & bd p@1 \\
\midrule
Small-4 & 11.1\% & 10.0\% & 0.51 & 3.42 \\
Small-5 & 13.9\% & 16.7\% & 0.46 & 3.43 \\
Small-6 & 16.7\% & 16.7\% & 0.42 & 2.85 \\
Small-7 & 19.4\% & 16.7\% & 0.42 & 3.40 \\
Medium-20 & 10.2\% & 1.7\% & 0.003 & 8.87 \\
Medium-40 & 20.4\% & 1.7\% & 0.80 & 8.52 \\
Medium-50 & 25.5\% & 0.0\% & --- & 8.45 \\
Medium-60 & 30.6\% & 6.7\% & 0.001 & 6.95 \\
Large-60 & 12.4\% & 3.3\% & 0.88 & 11.93 \\
Large-80 & 16.5\% & 1.7\% & 0.92 & 11.90 \\
Large-100 & 20.7\% & 1.7\% & 0.60 & 12.58 \\
Large-120 & 24.8\% & 3.3\% & 0.42 & 13.65 \\
Ultra-120 & 13.3\% & 0.0\% & --- & 17.28 \\
Ultra-140 & 15.6\% & 1.7\% & 0.80 & 15.97 \\
Ultra-160 & 17.8\% & 0.0\% & --- & 17.52 \\
Ultra-180 & 20.0\% & 0.0\% & --- & 18.92 \\
\bottomrule
\end{tabular}
\end{table}

\subsection{Predict: Accuracy by Environment Structure and Mission Type}
\label{apx:predict-analysis}

This section supplements Table~\ref{tab:predict-by-length} with breakdowns of Predict accuracy by the structure of the environment (Tables~\ref{tab:predict-structure} and \ref{tab:predict-objects}) and by the kind of mission the level poses (Tables~\ref{tab:predict-verb} and \ref{tab:predict-clauses}), together with the distribution of correct answers within each 3-sample cell (Table~\ref{tab:predict-consistency}). All accuracy figures are pass@1 over the same 320 cells. Two observations organise the tables. Accuracy is near ceiling for every model but Llama-4-Scout and Qwen2.5-72B on single-room levels and separates the models sharply once the grid becomes a nine-room maze. Within the maze levels, the presence of a locked door costs gpt-5.4 a further 7 points and Llama-4-Scout and Qwen2.5-72B 4 points each, while claude-opus-4-6, claude-sonnet-4-6, Kimi-K2.5 and Llama-4-Maverick all stay within 3 points of their unlocked-door scores (Table~\ref{tab:predict-structure}). Sample-level correlations between success and sequence length are $-0.54$ (claude-opus-4-6), $-0.38$ (claude-sonnet-4-6), $-0.51$ (Kimi-K2.5), $-0.71$ (gpt-5.4), $-0.76$ (Llama-4-Maverick), $-0.69$ (Llama-4-Scout) and $-0.76$ (Qwen2.5-72B). 
The coefficient is compressed where a model sits at ceiling over most of the range, so it should not be read as a ranking of how much length matters to each model.

\begin{table}[h]
\centering
\footnotesize
\setlength{\tabcolsep}{3pt} 
\caption{Predict: pass@1 success rate (\%) by environment structure. Single-room levels contain no doors, so the first column is also the no-door condition.}
\label{tab:predict-structure}
\begin{tabular}{l ccc}
\toprule
Model & 1 room ($n=120$) & 9 rooms, unlocked doors ($n=161$) & 9 rooms, locked door ($n=39$) \\
\midrule
claude-opus-4-6 & 100 & 87 & 90 \\
claude-sonnet-4-6 & 99 & 86 & 86 \\
Kimi-K2.5 & 100 & 97 & 96 \\
gpt-5.4 & 99 & 72 & 65 \\
Llama-4-Maverick & 96 & 50 & 52 \\
Llama-4-Scout & 83 & 39 & 35 \\
Qwen2.5-72B & 92 & 51 & 47 \\
\bottomrule
\end{tabular}
\end{table}

\begin{table}[h]
\centering
\footnotesize
\caption{Predict: pass@1 success rate (\%) by number of objects in the environment. Object count is confounded with room count: the 8-object bin is drawn mostly from single-room levels, which is why it scores above the 1-object bin.}
\label{tab:predict-objects}
\begin{tabular}{l cccc}
\toprule
Model & 1 ($n=40$) & 8 ($n=100$) & 18 ($n=118$) & 19--20 ($n=62$) \\
\midrule
claude-opus-4-6 & 92 & 100 & 88 & 87 \\
claude-sonnet-4-6 & 89 & 99 & 88 & 84 \\
Kimi-K2.5 & 99 & 100 & 97 & 97 \\
gpt-5.4 & 89 & 99 & 71 & 67 \\
Llama-4-Maverick & 68 & 95 & 52 & 53 \\
Llama-4-Scout & 57 & 83 & 41 & 35 \\
Qwen2.5-72B & 72 & 91 & 52 & 47 \\
\bottomrule
\end{tabular}
\end{table}

\begin{table}[h]
\centering
\footnotesize
\caption{Predict: pass@1 success rate (\%) by the verb of the mission. Unlike Decompose (Table~\ref{tab:decompose-verb}), Predict shows no comparable \texttt{PutNext} cliff. \texttt{PutNext} is the lowest column for every model, but only 2 to 11 points below the next-lowest verb.}
\label{tab:predict-verb}
\begin{tabular}{l cccc}
\toprule
Model & GoTo ($n=162$) & Open ($n=32$) & Pickup ($n=85$) & PutNext ($n=41$) \\
\midrule
claude-opus-4-6 & 93 & 95 & 93 & 87 \\
claude-sonnet-4-6 & 93 & 89 & 91 & 84 \\
Kimi-K2.5 & 99 & 97 & 99 & 95 \\
gpt-5.4 & 87 & 82 & 74 & 72 \\
Llama-4-Maverick & 71 & 68 & 64 & 60 \\
Llama-4-Scout & 62 & 48 & 51 & 42 \\
Qwen2.5-72B & 75 & 67 & 57 & 46 \\
\bottomrule
\end{tabular}
\end{table}

\begin{table}[h]
\centering
\footnotesize
\caption{Predict: pass@1 success rate (\%) by number of instruction clauses in the mission. Multi-clause missions carry longer action sequences, so this axis largely restates Table~\ref{tab:predict-by-length}.}
\label{tab:predict-clauses}
\begin{tabular}{l ccc}
\toprule
Model & 1 ($n=282$) & 2 ($n=24$) & 3+ ($n=14$) \\
\midrule
claude-opus-4-6 & 95 & 71 & 69 \\
claude-sonnet-4-6 & 93 & 79 & 60 \\
Kimi-K2.5 & 100 & 96 & 69 \\
gpt-5.4 & 87 & 49 & 29 \\
Llama-4-Maverick & 73 & 32 & 12 \\
Llama-4-Scout & 61 & 12 & 0 \\
Qwen2.5-72B & 73 & 15 & 7 \\
\bottomrule
\end{tabular}
\end{table}

\begin{table}[h]
\centering
\footnotesize
\caption{Predict: distribution of correct answers per 3-sample cell.}
\label{tab:predict-consistency}
\begin{tabular}{l rrrr}
\toprule
Model & 3/3 & 2/3 & 1/3 & 0/3 \\
\midrule
claude-opus-4-6 & 83.8\% & 10.3\% & 4.4\% & 1.6\% \\
claude-sonnet-4-6 & 80.9\% & 12.2\% & 5.0\% & 1.9\% \\
Kimi-K2.5 & 96.2\% & 2.2\% & 0.9\% & 0.6\% \\
gpt-5.4 & 69.1\% & 12.2\% & 12.2\% & 6.6\% \\
Llama-4-Maverick & 53.8\% & 14.1\% & 13.4\% & 18.8\% \\
Llama-4-Scout & 39.7\% & 16.2\% & 13.1\% & 30.9\% \\
Qwen2.5-72B & 51.6\% & 16.2\% & 10.0\% & 22.2\% \\
\bottomrule
\end{tabular}
\end{table}

\subsection{Decompose: What Precision Rate Tracks}
\label{apx:decompose-analysis}

The tables below break the Decompose results down by properties of the mission and of the environment, all at pass@1. Read together they say that Comprehension Rate falls far less than Precision Rate across almost every variable we can vary, while Precision Rate moves with all of them: most strongly with subgoal count, path length or room count, depending on the model (Table~\ref{tab:decompose-correlations}), and sharply with whether the mission can be planned in full before execution starts.

\begin{table}[h]
\centering
\footnotesize
\caption{Decompose: CR/PR/ACI at pass@1, by how many subgoals the oracle itself has to insert at runtime when starting from its own initial plan. This is a property of the mission, not of the model: it measures whether the oracle's own initial plan needs any runtime insertion. CR varies far less than PR across the three columns; PR collapses.}
\label{tab:decompose-static}
\begin{tabular}{l ccc}
\toprule
Model & oracle inserts 0 ($n=177$) & 1--2 ($n=74$) & 3+ ($n=69$) \\
\midrule
claude-opus-4-6 & 84/83/0.83 & 75/28/0.53 & 71/12/0.37 \\
claude-sonnet-4-6 & 89/88/0.88 & 91/41/0.68 & 87/9/0.46 \\
Kimi-K2.5 & 85/85/0.85 & 82/66/0.76 & 86/44/0.66 \\
gpt-5.4 & 83/80/0.81 & 89/30/0.58 & 82/1/0.37 \\
Llama-4-Maverick & 84/78/0.79 & 87/15/0.51 & 72/1/0.28 \\
Llama-4-Scout & 82/73/0.75 & 84/3/0.42 & 75/0/0.20 \\
Qwen2.5-72B & 77/64/0.67 & 78/4/0.35 & 70/0/0.18 \\
\bottomrule
\end{tabular}
\end{table}

\begin{table}[h]
\centering
\footnotesize
\caption{Decompose: CR/PR/ACI at pass@1, by the verb of the mission. The \texttt{PutNext} column is the one place in the benchmark where comprehension itself fails: four of seven models cannot produce a decomposition the oracle can complete. claude-opus-4-6's low \texttt{Open} figure is a convention artifact and is analysed in Appendix~\ref{apx:convention}.}
\label{tab:decompose-verb}
\begin{tabular}{l cccc}
\toprule
Model & GoTo ($n=162$) & Open ($n=32$) & Pickup ($n=85$) & PutNext ($n=41$) \\
\midrule
claude-opus-4-6 & 91/66/0.76 & 44/32/0.37 & 78/49/0.64 & 63/45/0.54 \\
claude-sonnet-4-6 & 99/73/0.85 & 81/55/0.68 & 98/54/0.78 & 33/23/0.27 \\
Kimi-K2.5 & 99/90/0.94 & 77/69/0.74 & 89/66/0.80 & 27/18/0.23 \\
gpt-5.4 & 98/67/0.80 & 95/54/0.71 & 93/44/0.69 & 4/2/0.03 \\
Llama-4-Maverick & 94/62/0.74 & 91/46/0.66 & 92/39/0.64 & 7/7/0.07 \\
Llama-4-Scout & 95/56/0.67 & 97/36/0.58 & 85/33/0.57 & 3/2/0.03 \\
Qwen2.5-72B & 90/52/0.63 & 91/18/0.44 & 78/29/0.47 & 3/2/0.02 \\
\bottomrule
\end{tabular}
\end{table}

\begin{table}[h]
\centering
\footnotesize
\caption{Decompose: CR/PR/ACI at pass@1, by the length in primitive actions of the oracle's execution path. This is the axis that connects Decompose to the Plan results of Section~\ref{sec:plan}.}
\label{tab:decompose-pathlen}
\begin{tabular}{l cccc}
\toprule
Model & $\leq$10 ($n=159$) & 11--20 ($n=54$) & 21--35 ($n=44$) & 36+ ($n=63$) \\
\midrule
claude-opus-4-6 & 88/86/0.87 & 65/36/0.53 & 72/27/0.45 & 71/13/0.39 \\
claude-sonnet-4-6 & 95/93/0.94 & 78/48/0.66 & 87/30/0.58 & 84/8/0.43 \\
Kimi-K2.5 & 92/92/0.92 & 69/56/0.64 & 85/67/0.77 & 80/39/0.60 \\
gpt-5.4 & 91/88/0.89 & 69/32/0.50 & 89/17/0.48 & 75/2/0.35 \\
Llama-4-Maverick & 91/84/0.86 & 72/27/0.50 & 83/5/0.39 & 66/1/0.25 \\
Llama-4-Scout & 91/79/0.82 & 60/12/0.39 & 82/0/0.27 & 73/1/0.22 \\
Qwen2.5-72B & 85/68/0.72 & 64/12/0.36 & 73/0/0.23 & 67/1/0.19 \\
\bottomrule
\end{tabular}
\end{table}

\begin{table}[h]
\centering
\footnotesize
\setlength{\tabcolsep}{3pt} 
\caption{Decompose: CR/PR at pass@1, by environment structure. On single-room levels the oracle's own plan never needs a runtime insertion, and CR and PR coincide for every model except Qwen2.5-72B; PR and ACI are informative only in the multi-room columns.}
\label{tab:decompose-structure}
\begin{tabular}{l cccc}
\toprule
Model & 1 room ($n=120$) & 9 rooms ($n=200$) & unlocked doors ($n=161$) & locked door ($n=39$) \\
\midrule
claude-opus-4-6 & 93/93 & 71/33 & 72/34 & 63/25 \\
claude-sonnet-4-6 & 90/90 & 88/42 & 92/43 & 68/35 \\
Kimi-K2.5 & 84/84 & 85/65 & 89/70 & 70/43 \\
gpt-5.4 & 79/79 & 87/35 & 93/36 & 60/29 \\
Llama-4-Maverick & 84/84 & 81/25 & 88/26 & 53/21 \\
Llama-4-Scout & 79/79 & 82/19 & 87/19 & 61/20 \\
Qwen2.5-72B & 78/76 & 75/12 & 81/13 & 52/11 \\
\bottomrule
\end{tabular}
\end{table}

\begin{table}[h]
\centering
\footnotesize
\caption{Decompose: CR/PR/ACI at pass@1 within the 1--2 oracle-subgoal-count band only ($n=131$ cells, the 1--2 column of Table~\ref{tab:decompose-by-count}), split by object count. Holding decomposition difficulty fixed, clutter still costs precision for every model while leaving comprehension largely intact for every model except claude-opus-4-6.}
\label{tab:decompose-clutter}
\begin{tabular}{l cc}
\toprule
Model & few objects ($n=86$) & many objects ($n=45$) \\
\midrule
claude-opus-4-6 & 95/90/0.92 & 65/54/0.59 \\
claude-sonnet-4-6 & 99/97/0.98 & 90/82/0.86 \\
Kimi-K2.5 & 100/98/0.99 & 92/88/0.90 \\
gpt-5.4 & 99/95/0.97 & 96/80/0.85 \\
Llama-4-Maverick & 100/96/0.98 & 88/66/0.74 \\
Llama-4-Scout & 96/90/0.92 & 96/56/0.70 \\
Qwen2.5-72B & 95/86/0.89 & 87/35/0.51 \\
\bottomrule
\end{tabular}
\end{table}

\begin{table}[h]
\centering
\footnotesize
\caption{Decompose: quality of the decompositions that did reach the goal, conditioning on CR${}=1$. Even among successful episodes the models leave the oracle between 0.6 and 1.7 subgoals to supply, and on long missions the share of the work they contribute themselves separates them widely.}
\label{tab:decompose-quality}
\begin{tabular}{l rrrr}
\toprule
Model & PR $\mid$ CR${}=1$ & ACI $\mid$ CR${}=1$ & mean inserted & ACI $\mid$ CR${}=1$, 7+ subgoals \\
\midrule
claude-opus-4-6 & 70\% & 0.84 & 1.0 & 0.54 \\
claude-sonnet-4-6 & 68\% & 0.84 & 1.1 & 0.49 \\
Kimi-K2.5 & 85\% & 0.93 & 0.6 & 0.70 \\
gpt-5.4 & 61\% & 0.79 & 1.2 & 0.43 \\
Llama-4-Maverick & 57\% & 0.75 & 1.3 & 0.38 \\
Llama-4-Scout & 51\% & 0.69 & 1.6 & 0.32 \\
Qwen2.5-72B & 48\% & 0.64 & 1.7 & 0.28 \\
\bottomrule
\end{tabular}
\end{table}

\begin{table}[h]
\centering
\footnotesize
\caption{Decompose: CR/PR at pass@1 by number of instruction clauses, excluding \texttt{PutNext} missions. Half of the raw 2-clause bin is \texttt{PutNext}, which inverts the CR ordering; with those missions removed CR is flat and PR falls steeply. Mean oracle subgoal counts across the three bins are 3.0, 7.7 and 11.4, so this axis is largely a restatement of mission length.}
\label{tab:decompose-clauses}
\begin{tabular}{l ccc}
\toprule
Model & 1 clause ($n=255$) & 2 clauses ($n=12$) & 3+ clauses ($n=12$) \\
\midrule
claude-opus-4-6 & 82/61 & 75/22 & 81/3\\
claude-sonnet-4-6 & 96/70 & 100/31 & 100/6 \\
Kimi-K2.5 & 94/84 & 92/61 & 89/6 \\
gpt-5.4 & 96/64 & 94/11 & 89/0 \\
Llama-4-Maverick & 93/58 & 86/0 & 92/0 \\
Llama-4-Scout & 92/51 & 86/3 & 100/0 \\
Qwen2.5-72B & 86/45 & 86/0 & 94/0 \\
\bottomrule
\end{tabular}
\end{table}

\begin{table}[h]
\centering
\footnotesize
\setlength{\tabcolsep}{3pt} 
\caption{Decompose: cell-level correlation of PR (pass@1) with each difficulty variable. Mission-intrinsic variables (subgoal count, path length) dominate for Kimi-K2.5, claude-sonnet-4-6 and gpt-5.4. For claude-opus-4-6, Llama-4-Scout and Qwen2.5-72B the environment-scale variables are the stronger predictors, and in the case of claude-opus-4-6 this partly reflects the convention artifact of Appendix~\ref{apx:convention}, since multi-room levels are also the levels with doors to open.}
\label{tab:decompose-correlations}
\resizebox{\linewidth}{!}{%
\begin{tabular}{l cccccccc}
\toprule
Model & oracle subgoals & initial plan size & oracle inserts & path length & clauses & objects & doors & rooms \\
\midrule
claude-opus-4-6 & $-0.58$ & $-0.32$ & $-0.59$ & $-0.61$ & $-0.34$ & $-0.49$ & $-0.64$ & $-0.68$ \\
claude-sonnet-4-6 & $-0.69$ & $-0.46$ & $-0.64$ & $-0.70$ & $-0.39$ & $-0.41$ & $-0.51$ & $-0.54$ \\
Kimi-K2.5 & $-0.69$ & $-0.67$ & $-0.50$ & $-0.55$ & $-0.48$ & $-0.32$ & $-0.23$ & $-0.25$ \\
gpt-5.4 & $-0.65$ & $-0.48$ & $-0.58$ & $-0.64$ & $-0.34$ & $-0.35$ & $-0.43$ & $-0.46$ \\
Llama-4-Maverick & $-0.63$ & $-0.44$ & $-0.58$ & $-0.63$ & $-0.33$ & $-0.46$ & $-0.59$ & $-0.62$ \\
Llama-4-Scout & $-0.59$ & $-0.42$ & $-0.54$ & $-0.59$ & $-0.29$ & $-0.43$ & $-0.59$ & $-0.63$ \\
Qwen2.5-72B & $-0.55$ & $-0.41$ & $-0.50$ & $-0.54$ & $-0.28$ & $-0.51$ & $-0.67$ & $-0.70$ \\
\bottomrule
\end{tabular}}
\end{table}

\begin{table}[h]
\centering
\footnotesize
\caption{Decompose: distribution of successful samples per 3-sample cell, for CR and for PR. The PR column shows the same two-component structure as Plan: a large deterministic-failure mass (0/3) that grows as models weaken, plus a variance component.}
\label{tab:decompose-consistency}
\begin{tabular}{l rrrr rrrr}
\toprule
& \multicolumn{4}{c}{CR} & \multicolumn{4}{c}{PR} \\
\cmidrule(lr){2-5}\cmidrule(lr){6-9}
Model & 3/3 & 2/3 & 1/3 & 0/3 & 3/3 & 2/3 & 1/3 & 0/3 \\
\midrule
claude-opus-4-6 & 61\% & 20\% & 13\% & 6\% & 41\% & 14\% & 13\% & 31\% \\
claude-sonnet-4-6 & 81\% & 10\% & 3\% & 6\% & 48\% & 12\% & 10\% & 29\% \\
Kimi-K2.5 & 74\% & 13\% & 7\% & 6\% & 58\% & 15\% & 12\% & 15\% \\
gpt-5.4 & 80\% & 3\% & 4\% & 12\% & 45\% & 6\% & 8\% & 41\% \\
Llama-4-Maverick & 73\% & 11\% & 5\% & 11\% & 39\% & 7\% & 10\% & 44\% \\
Llama-4-Scout & 71\% & 13\% & 3\% & 13\% & 34\% & 8\% & 6\% & 52\% \\
Qwen2.5-72B & 62\% & 16\% & 8\% & 13\% & 27\% & 9\% & 9\% & 55\% \\
\bottomrule
\end{tabular}
\end{table}

\subsection{\texorpdfstring{The Coordinate Convention in \texttt{GoNextToSubgoal}}{The Coordinate Convention in GoNextToSubgoal}}
\label{apx:convention}

Our prompt describes \texttt{GoNextToSubgoal, (x,y)} as an instruction to go next to the cell $(x,y)$. The intended argument is the target object's own coordinate, and the executor then walks the agent to a free adjacent cell facing it. A model can also read the instruction as asking for the cell the agent should finish on, and pass an adjacent coordinate instead. The executor then places the agent next to that cell, one square too far from the object, and the \texttt{OpenSubgoal} or \texttt{PickupSubgoal} that follows has nothing in front of it. The following response is typical of the second reading:

\begin{tcolorbox}[colback=verylightgray, title={A sample that applies ``next to'' twice}]
\ttfamily\footnotesize
The green door is at (19, 7). [...] The agent can approach from (19, 8) which is south of the door [...] Let me go next to the door and open it.

\medskip
(GoNextToSubgoal, (19, 8))\\
(OpenSubgoal)
\end{tcolorbox}

Tables~\ref{tab:convention-open} and \ref{tab:convention-all} give the frequency of each reading and the completion rate conditional on it. 
The effect is large for every model that uses the adjacent form, though it varies widely, from 11 points for Llama-4-Maverick to 52 for Llama-4-Scout.
What distinguishes claude-opus-4-6 is only how often it does so. Its aggregate Open score follows arithmetically from the mix, since $0.45 \times 93\% + 0.55 \times 6\% \approx 45\%$, within rounding of the 44\% that Table~\ref{tab:decompose-verb} reports. The same tendency appears elsewhere in its output: before a \texttt{PickupSubgoal} it passes an adjacent cell in 21\% of samples against 5\% for claude-sonnet-4-6, and on plain \texttt{GoTo} targets in 12\% against 3\%. We report raw scores everywhere in the paper. 

\begin{table}[h]
\centering
\footnotesize
\caption{Open level: how often each model passes the door's own coordinate rather than an adjacent cell, and the pass@1 completion rate (\%) conditional on each choice. Shares do not always sum to 100 because a few samples pass a coordinate that is neither. Cells marked n/a are not reported.}
\label{tab:convention-open}
\begin{tabular}{l rr cc}
\toprule
Model & uses object cell & uses adjacent cell & CR when object cell & CR when adjacent \\
\midrule
claude-opus-4-6 & 45\% & 55\% & 93 ($n=27$) & 6 ($n=33$) \\
claude-sonnet-4-6 & 82\% & 18\% & 98 ($n=49$) & 0 ($n=11$) \\
Kimi-K2.5 & 84\% & 16\% & 94 ($n=48$) & 0 ($n=9$) \\
gpt-5.4 & 100\% & 0\% & 96 ($n=57$) & n/a \\
Llama-4-Maverick & 90\% & 8\% & n/a & n/a \\
Llama-4-Scout & 95\% & 5\% & 100 ($n=57$) & 67 ($n=3$) \\
Qwen2.5-72B & 75\% & 23\% & n/a & n/a \\
\bottomrule
\end{tabular}
\end{table}

\begin{table}[h]
\centering
\footnotesize
\caption{All Decompose levels: the same breakdown for every \texttt{GoNextToSubgoal} that immediately precedes an interaction subgoal, with pass@1 completion rate (\%) conditional on the convention used. Restricted to the documented convention, claude-opus-4-6 has the highest comprehension rate of the seven models.}
\label{tab:convention-all}
\begin{tabular}{l rr cc}
\toprule
Model & uses object cell & uses adjacent cell & CR when object cell & CR when adjacent \\
\midrule
claude-opus-4-6 & 59\% & 41\% & 89 & 50 \\
claude-sonnet-4-6 & 91\% & 9\% & 86 & 57 \\
Kimi-K2.5 & 88\% & 12\% & 84 & 34 \\
gpt-5.4 & 100\% & 0\% & 78 & n/a \\
Llama-4-Maverick & 89\% & 8\% & 75 & 64 \\
Llama-4-Scout & 96\% & 4\% & 75 & 23 \\
Qwen2.5-72B & 80\% & 17\% & 74 & 48 \\
\bottomrule
\end{tabular}
\end{table}

\subsection{Missing Subgoals in Long Decompositions}
\label{apx:decompose-audit}

Precision Rate is zero for all seven models on all 19 instances whose oracle decomposition runs to ten or more subgoals, so the 399 samples in that band cover every case of this failure. We re-ran those episodes with the executor instrumented to record every subgoal it injected and the point at which it did so. In two samples the executor reached an unrecoverable state and raised an exception instead of terminating; the remaining 397 ran to completion, and all 399 are included in the counts below.

\begin{table}[h]
\centering
\footnotesize
\caption{Decompose, missions of 10+ oracle subgoals: what the executor had to inject. The middle columns give the type of the \emph{first} injection, which is where the model's plan first became unexecutable; 
rows do not sum to 100\% because a small number of episodes ran out of plan before any injection was needed, and because one sample's first injection is a \texttt{Close}, which is omitted from the table due to its rarity.}
\label{tab:audit-injections}
\begin{tabular}{lrr rrr rr}
\toprule
& & mean & \multicolumn{3}{c}{first injection is} & \multicolumn{2}{c}{any injected} \\
\cmidrule(lr){4-6}\cmidrule(lr){7-8}
Model & $n$ & injected & \texttt{Open} & \texttt{Pickup}/\texttt{Drop} & \texttt{GoNextTo} & \texttt{Open} & \texttt{Pickup}/\texttt{Drop} \\
\midrule
claude-opus-4-6   & 57 &  7.6 & 72\% & 21\% &  5\% &  82\% & 74\% \\
claude-sonnet-4-6 & 57 &  9.9 & 70\% & 11\% & 18\% &  89\% & 91\% \\
Kimi-K2.5         & 57 &  5.8 & 46\% & 32\% & 11\% &  51\% & 72\% \\
gpt-5.4           & 57 &  8.5 & 84\% &  9\% &  7\% &  96\% & 82\% \\
Llama-4-Maverick  & 57 &  9.1 & 89\% &  0\% &  9\% & 100\% & 84\% \\
Llama-4-Scout     & 57 &  9.0 & 81\% &  0\% & 19\% &  96\% & 84\% \\
Qwen2.5-72B       & 57 &  9.4 & 81\% &  0\% & 19\% &  98\% & 84\% \\
\bottomrule
\end{tabular}
\end{table}

\begin{table}[h]
\centering
\small
\setlength{\tabcolsep}{3pt} 
\caption{Decompose, missions of 10+ oracle subgoals: length of the plans models produce against the oracle's.}
\label{tab:audit-length}
\begin{tabular}{lrrrr}
\toprule
Model & mean LLM subgoals & mean oracle subgoals & ratio & plans shorter than oracle \\
\midrule
claude-opus-4-6   & 12.0 & 13.4 & 0.90 &  54\% \\
claude-sonnet-4-6 &  9.1 & 13.4 & 0.68 &  81\% \\
Kimi-K2.5         & 10.4 & 13.4 & 0.77 &  68\% \\
gpt-5.4           &  6.7 & 13.4 & 0.50 & 100\% \\
Llama-4-Maverick  &  6.8 & 13.4 & 0.51 & 100\% \\
Llama-4-Scout     &  5.5 & 13.4 & 0.41 &  98\% \\
Qwen2.5-72B       &  5.5 & 13.4 & 0.41 & 100\% \\
\bottomrule
\end{tabular}
\end{table}

\textbf{Example: Kimi-K2.5 on BossLevel, seed 31.} It has 20 oracle subgoals, tied for the most in the band, and eight doors, one of them locked. The mission is \emph{put the box behind you next to the red door and pick up a yellow ball}, and the oracle needs seven \texttt{GoNextTo}, five \texttt{Open}, four \texttt{Pickup} and four \texttt{Drop} subgoals. The model produced six subgoals and no door openings at all:

\begin{tcolorbox}[colback=verylightgray, title={Kimi-K2.5, BossLevel seed 31: the complete plan}]
\ttfamily\footnotesize
(GoNextToSubgoal, (4,16))\\
(PickupSubgoal)\\
(GoNextToSubgoal, (14,15))\\
(DropSubgoal)\\
(GoNextToSubgoal, (16,17))\\
(PickupSubgoal)
\end{tcolorbox}

The executor injected 17 subgoals, opening with three consecutive door-openings before any of the model's own subgoals could run, and the mission was still not completed when the episode ended. Its injections totalled six \texttt{Open}, six \texttt{GoNextTo}, three \texttt{Drop} and two \texttt{Pickup}. The plan is a reasonable outline of the goal, but it contains none of the traversal the mission requires.

\textbf{Example: gpt-5.4 on BossLevel, seed 67.} The instance has twenty oracle subgoals and nine doors. The mission is \emph{pick up a green key after you open a red door and go to the blue door}. The episode succeeded on CR and scored an ACI of 0.45 because the executor supplied 11 of the subgoals needed to complete it.

\begin{tcolorbox}[colback=verylightgray, title={gpt-5.4, BossLevel seed 67: the complete plan}]
\ttfamily\footnotesize
(GoNextToSubgoal, (12,14))\\
(OpenSubgoal)\\
(GoNextToSubgoal, (4,14))\\
(GoNextToSubgoal, (5,9))\\
(PickupSubgoal)
\end{tcolorbox}

The single \texttt{OpenSubgoal} is for the red door named in the mission. The executor then had to inject three more \texttt{Open} subgoals for doors the route passes through, along with six \texttt{GoNextTo}, one \texttt{Pickup} and one \texttt{Drop}. The same pattern appears across the band. Models open the doors named in the mission and omit the other doors on the route.

\begin{table}[h]
\centering
\small
\caption{Decompose: subgoal coverage on missions of ten or more oracle subgoals, pooled over all 399 samples (19 instances $\times$ 7 models $\times$ 3 samples). Coverage is the fraction of the subgoals the episode required that the model supplied itself, that is $1-\text{injected}/\text{required}$.}
\label{tab:decompose-audit}
\begin{tabular}{lrrrr}
\toprule
Subgoal type & emitted & required & injected & coverage \\
\midrule
\texttt{GoNextTo} (navigation) & 4.32 & 4.59 & 2.76 & 40\% \\
\texttt{Open} (door)           & 1.68 & 4.37 & 3.36 & 23\% \\
\texttt{Pickup}                & 1.48 & 2.26 & 0.96 & 57\% \\
\texttt{Drop} (unblocking)     & 0.54 & 2.20 & 1.35 & 39\% \\
\bottomrule
\end{tabular}
\end{table}

\subsection{The L-Route Strategy}
\label{apx:lroute}

This section supports Section~\ref{sec:plan-failure} with a worked trace, the verification counts behind it, and the detour breakdown the strategy predicts. The four models with an extended-reasoning control ran with it at its default setting (Section~\ref{sec:results}), and claude-opus-4-6 and claude-sonnet-4-6 write their working step by step in the body of the response and the procedure is visible without instrumentation. One successful Ultra response reads:

\begin{quote}\small
I need to go right (east) by 11 cells (from $x{=}4$ to $x{=}15$). I need to go down (south) by 14 cells\dots{} Wait, I need to check for obstacles along the path. Path east at $y{=}4$ from $x{=}4$: $x=5,6,\dots,15$\dots{} I don't see any objects blocking this row. Path south at $x{=}15$ from $y{=}4$ to $y{=}17$: $(15,5)$ clear, $(15,6)$ clear, \dots{} $(15,9)$, ball at $(15,9)$! Blocked. I need to reroute\dots{} Let me try $x{=}12$ south from $y{=}4$\dots{} Clear!
\end{quote}

The pattern is the same across instances. The model computes the displacement to the target, proposes an L-shaped corridor of two straight runs joined by a turn, scans the object list for anything lying on that corridor, and shifts the corridor by a row or a column when the scan reports a blocker. It commits on the first clean scan. In the traces we examined the branching factor of this search is three or four. It is systematic, although it falls well short of a shortest-path algorithm. Successful plans of claude-opus-4-6, claude-sonnet-4-6, Kimi-K2.5, gpt-5.4 and Llama-4-Maverick run 1.28 to 1.37 times the free-space lower bound, which is an L-route plus its two or three turns. This is consistent with the efficiency column of Table~\ref{tab:plan}, which uses a different reference. That column compares a plan against the OmniBot's own path, which must itself detour around the same obstacles. The 1.28 to 1.37 figure is taken against the free-space lower bound, which ignores them.

\begin{table}[h]
\centering
\small
\caption{Plan: obstacle checking in the response text, on the Small and Ultra tracks. Mean occurrences per response of \emph{check} and of \emph{block}, and the share of responses containing at least one occurrence of either. Keyword counts are a coarse proxy for the underlying procedure. They are reported for the five models whose responses carry some working text: Kimi-K2.5 returns little text beyond its answer, and Qwen2.5-72B's responses are dominated by repetition (Appendix~\ref{apx:effort}).}
\label{tab:plan-verbalisation}
\begin{tabular}{l cc cc cc}
\toprule
& \multicolumn{2}{c}{``check'' per response} & \multicolumn{2}{c}{``block'' per response} & \multicolumn{2}{c}{responses that check} \\
\cmidrule(lr){2-3}\cmidrule(lr){4-5}\cmidrule(lr){6-7}
Model & Small & Ultra & Small & Ultra & Small & Ultra \\
\midrule
claude-opus-4-6   & 0.6 & 5.6 & 0.8 & 4.6 & 41\% & 100\% \\
claude-sonnet-4-6 & 0.3 & 6.4 & 1.3 & 7.5 & 24\% &  99\% \\
gpt-5.4           & 0.0 & 0.7 & 0.3 & 2.9 &  1\% &  38\% \\
Llama-4-Maverick  & 0.0 & 0.0 & 0.0 & 0.0 &  0\% &   2\% \\
Llama-4-Scout     & 0.1 & 0.1 & 0.3 & 0.2 &  8\% &   9\% \\
\bottomrule
\end{tabular}
\end{table}

Models check for obstacles more often on more cluttered grids (Table~\ref{tab:plan-verbalisation}). The share of claude-opus-4-6 responses that check an obstacle at all rises from 41\% on Small to 100\% on Ultra, and of claude-sonnet-4-6 responses from 24\% to 99\%, with the mean number of mentions rising by roughly an order of magnitude over the same range. Checking also separates the model tiers. Llama-4-Maverick records almost no occurrences at any size and Llama-4-Scout no increase with size, and these are the two models that fail even on the instances a direct route solves (32\% and 13\%, Table~\ref{tab:plan-by-detour}). The strong models do use the obstacle list, but their procedure is a linear pass over an unordered list of 120 to 180 objects, written out in the response text, that keeps no record of what an earlier pass has already ruled out.

\textbf{Completion by required detour.} The corridor procedure suggests which instances a model solves. Let the \emph{detour} an instance requires be its oracle plan length minus the free-space lower bound, so that a detour of zero means a direct corridor reaches the goal and a large detour means no such corridor exists. Completion falls sharply with required detour (Table~\ref{tab:plan-by-detour}). claude-opus-4-6 solves 99\% of instances requiring a detour of at most three actions and 5\% of those requiring seventeen or more, claude-sonnet-4-6 96\% and 5\%, and Kimi-K2.5, whose profile is flatter, 88\% and 29\%.

\begin{table}[h]
\centering
\small
\caption{Plan: pass@1 completion rate (\%) by the detour the optimal solution requires, in actions, with all grid sizes pooled. Detour is the oracle plan length minus the free-space lower bound; a detour of at most three counts as near-direct. Qwen2.5-72B is at floor across all bins and is omitted.}
\label{tab:plan-by-detour}
\begin{tabular}{l rrrr}
\toprule
Model & 0--3 & 4--8 & 9--16 & 17+ \\
\midrule
Kimi-K2.5         & 88\% & 69\% & 39\% & 29\% \\
claude-opus-4-6   & 99\% & 69\% & 17\% &  5\% \\
claude-sonnet-4-6 & 96\% & 56\% & 18\% &  5\% \\
gpt-5.4           & 53\% & 10\% &  6\% &  0\% \\
Llama-4-Maverick  & 32\% &  1\% &  0\% &  0\% \\
Llama-4-Scout     & 13\% &  0\% &  0\% &  0\% \\
\bottomrule
\end{tabular}
\end{table}

The supply of such instances changes with grid size. The mean detour a solution requires rises from 2.5 actions on Small to 5.2 on Medium and then flattens, at 5.0 on Large and 5.0 on Ultra, while the share of instances that a near-direct route solves falls throughout, from 79\% on Small to 49\%, 45\% and 32\%. Larger grids therefore contain fewer of the instances this procedure handles, in addition to requiring longer paths, and part of the size effect in Table~\ref{tab:plan-by-size} comes from this change in composition. It is also consistent with the plans models do return staying near-optimal at every size, although a substantial share of the strong models' successes on the larger grids are on instances that no near-direct route solves (Table~\ref{tab:plan-by-detour}). Required detour and oracle plan length are correlated by construction, so these numbers are a refinement of the horizon result of Table~\ref{tab:plan-by-length}. Separating the two would require stratifying detour at fixed oracle length.

\textbf{Obstacle checking in failed responses.} On the Ultra track claude-opus-4-6's failed responses contain more obstacle checking than its successful ones, 7.9 mentions of \emph{block} against 2.5, in 3{,}598 characters against 2{,}714. A representative failed response scans row $y{=}14$ and finds a blocker at $x{=}13$, tries $y{=}13$ and finds one at $x{=}24$, tries $y{=}15$ and finds five, returns to $y{=}14$, attempts a jog around the first blocker, meets a ball, and commits to a route whose remaining segments it never scans. Nothing in these responses maintains a frontier, a set of positions already visited, or a record of which corridors have been eliminated, so when no simple corridor is clear, the procedure has no stored state to backtrack to and ends on a route it has not verified. This is consistent with the execution profile of Table~\ref{tab:plan-blocks}: the plan runs as intended for the first 20 to 55\% of its actions, then meets the distractor the scan never reached, and every action after that fires from a pose the plan was not written against.

Two limits apply to this evidence. First, occurrence counts for \emph{check} and \emph{block} are a lexical proxy for verification rather than a measurement of it. Second, they are available only for the models whose responses carry some working text, which excludes Kimi-K2.5 and Qwen2.5-72B (Appendix~\ref{apx:effort}). Third, the procedural account is drawn from one response read in full, and checked against further responses with model assistance.

On this account, the models verify candidate routes but keep no record of the routes they have already ruled out, and the failure is due to this missing record. The account predicts that supplying such a record should reduce the collapse, and that a prompt asking only for more careful checking should not, since the failing responses already check more than the successful ones. We have not tested this.

\subsection{Trace of a Failed Plan}
\label{apx:plan-trace}

\begin{table}[h]
\centering
\small
\setlength{\tabcolsep}{3pt} 
\caption{Plan: collisions in failed plans on the Ultra ($32\times32$) track. A block is a \texttt{forward} the environment refuses because the target cell is occupied. Position of the first block is expressed as a fraction of the plan's length. Llama-4-Maverick and Llama-4-Scout are omitted.}
\label{tab:plan-blocks}
\resizebox{\linewidth}{!}{%
\begin{tabular}{lrrrrr}
\toprule
Model & failed samples & mean blocks & plans with $\geq 1$ block & first block at & mean plan length \\
\midrule
claude-opus-4-6   &  94 & 13.7 & 95\% & 44\% & 37 \\
claude-sonnet-4-6 & 118 & 32.7 & 96\% & 43\% & 97 \\
Kimi-K2.5         &  64 &  7.0 & 69\% & 55\% & 31 \\
gpt-5.4           & 186 & 12.0 & 88\% & 39\% & 26 \\
Qwen2.5-72B       & 239 & 482.0 & 97\% & 20\% & 1902 \\
\bottomrule
\end{tabular}}
\end{table}

Table~\ref{tab:trace} follows one failed Ultra episode action by action. The grid is $32\times32$ with 160 distractors, the agent starts at $(16,20)$ facing south, and the red ball is at $(1,18)$. The oracle solves it in 20 actions; Kimi-K2.5 emitted 23 and finished eight cells away.

\begin{table}[h]
\centering
\footnotesize
\caption{Kimi-K2.5 on the Ultra track, seed 43, 160 distractors, under the zero-shot structured prompt used for all reported results. Every action up to the twelfth executes as the model intended. The thirteenth is refused, and from that point the agent's pose no longer matches the one the plan was written against.}
\label{tab:trace}
\begin{tabular}{r l l l l}
\toprule
\# & action & from & facing & outcome \\
\midrule
1 & right & $(16,20)$ & south & now facing west \\
2--6 & forward $\times 5$ & $(16,20)$ & west & moves to $(11,20)$, all five succeed \\
7 & right & $(11,20)$ & west & now facing north \\
8--10 & forward $\times 3$ & $(11,20)$ & north & moves to $(11,17)$, all three succeed \\
11 & left & $(11,17)$ & north & now facing west \\
12 & forward & $(11,17)$ & west & moves to $(10,17)$ \\
13 & forward & $(10,17)$ & west & \textbf{blocked by a grey ball at $(9,17)$} \\
14--21 & forward $\times 8$ & $(10,17)$ & west & blocked by the same grey ball, eight times \\
22 & left & $(10,17)$ & west & now facing south \\
23 & forward & $(10,17)$ & south & moves to $(10,18)$; episode ends \\
\bottomrule
\end{tabular}
\end{table}

Nine of the twenty-three actions are the same refused \texttt{forward}, driven against a grey ball at $(9,17)$ that the prompt lists with its coordinates. Because the sequence is executed open-loop, nothing tells the model the cell is occupied, and it repeats the action eight further times before the plan turns elsewhere. The first divergence falls at action 13 of 23, 57\% of the way through, close to the 55\% figure for this model (Table~\ref{tab:plan-blocks}). The episode ends at $(10,18)$, eight cells from the target.

\subsection{Response Length and Latency Against Problem Size}
\label{apx:effort}

\begin{table}[h]
\centering
\small
\caption{Plan: response length in characters divided by the number of actions in the oracle solution, by grid size. Lower means less output per unit of required solution. Response length is a coarse proxy for computation; the figures are comparable across grid sizes within a model, not between models.}
\label{tab:plan-effort}
\begin{tabular}{lrrrr}
\toprule
Model & Small & Medium & Large & Ultra \\
\midrule
claude-opus-4-6   &  291 &  231 &  191 &  161 \\
claude-sonnet-4-6 &  345 &  272 &  237 &  187 \\
Kimi-K2.5         &   20 &   17 &   13 &   11 \\
gpt-5.4           &   78 &   70 &   61 &   57 \\
Llama-4-Maverick  &  340 &  187 &  164 &  117 \\
Llama-4-Scout     &  348 &  193 &  134 &   91 \\
Qwen2.5-72B       & 1075 & 2655 & 1416 &  945 \\
\bottomrule
\end{tabular}
\end{table}

We did not record token counts, so the figures below use response length in characters and per-call wall-clock latency as proxies for how much work a model puts into an instance. Both proxies are crude. Characters are not tokens, latency depends on serving conditions we do not control, and a model can think for a long time without producing much text. They are, however, the only effort signals in our logs, and they agree clearly on the direction of the effect. We therefore compare these figures only within a model, across grid sizes.

\begin{table}[h]
\centering
\footnotesize
\caption{Plan: response length, latency and completion by grid size. The prompt grows by a factor of 3.44 from Small to Ultra for every model, since all use the same formatter, and the oracle solution grows from roughly 7 actions to 24. Latency by grid size was not retained for the three open non-reasoning models.}
\label{tab:effort-full}
\begin{tabular}{l l rrrr r}
\toprule
Model & measure & Small & Medium & Large & Ultra & Ultra/Small \\
\midrule
\multirow{4}{*}{Kimi-K2.5}
 & response chars & 116 & 181 & 208 & 252 & 2.17$\times$ \\
 & latency (s) & 223 & 265 & 293 & 308 & 1.38$\times$ \\
 & prompt chars & 2283 & 3676 & 5466 & 7854 & 3.44$\times$ \\
 & SR pass@1 & 89.2\% & 77.9\% & 70.8\% & 61.7\% & 0.69$\times$ \\
\midrule
\multirow{3}{*}{claude-opus-4-6}
 & response chars & 1648 & 2553 & 2818 & 3060 & 1.86$\times$ \\
 & latency (s) & 12 & 20 & 22 & 25 & 2.01$\times$ \\
 & SR pass@1 & 100\% & 79\% & 70\% & 61\% & 0.61$\times$ \\
\midrule
\multirow{3}{*}{claude-sonnet-4-6}
 & response chars & 1929 & 3157 & 3531 & 3773 & 1.96$\times$ \\
 & latency (s) & 13 & 21 & 25 & 24 & 1.93$\times$ \\
 & SR pass@1 & 97\% & 72\% & 64\% & 51\% & 0.52$\times$ \\
\midrule
\multirow{3}{*}{gpt-5.4}
 & response chars & 442 & 888 & 1044 & 1247 & 2.82$\times$ \\
 & latency (s) & 4 & 6 & 7 & 8 & 2.08$\times$ \\
 & SR pass@1 & 52\% & 28\% & 25\% & 22\% & 0.44$\times$ \\
\midrule
\multirow{2}{*}{Llama-4-Maverick}
 & response chars & 1853 & 1854 & 2186 & 2145 & 1.16$\times$ \\
 & SR pass@1 & 35\% & 17\% & 9\% & 6\% & 0.16$\times$ \\
\midrule
\multirow{2}{*}{Llama-4-Scout}
 & response chars & 1882 & 1805 & 1671 & 1558 & 0.83$\times$ \\
 & SR pass@1 & 18\% & 6\% & 3\% & 1\% & 0.05$\times$ \\
\midrule
\multirow{2}{*}{Qwen2.5-72B}
 & response chars & 5645 & 22489 & 18659 & 17495 & 3.10$\times$ \\
 & SR pass@1 & 15\% & 2\% & 2\% & 0\% & 0.03$\times$ \\
\bottomrule
\end{tabular}
\end{table}

Absolute output grows for six of the seven models, so the models do not ignore problem size entirely. However, it grows more slowly than the solution does, as the per-unit figures of Table~\ref{tab:plan-effort} show, and for Llama-4-Scout and Llama-4-Maverick it changes little (0.83 and 1.16 times) while their completion rates fall by factors of 20 and 6. For Qwen2.5-72B, output rises from 5645 characters on Small to 22489 on Medium and then falls again, a profile consistent with the repetition and looping we observe in its transcripts rather than with sustained deliberation.

\begin{table}[h]
\centering
\small
\caption{Plan, Ultra track: response length and latency split by whether the episode succeeded.}
\label{tab:effort-outcome}
\begin{tabular}{lrrrr}
\toprule
Model & chars when solved & chars when failed & latency solved & latency failed \\
\midrule
claude-opus-4-6   & 2714 & 3598 & 21s & 31s \\
claude-sonnet-4-6 & 3555 & 3999 & 24s & 25s \\
Kimi-K2.5         &  225 &  315 & 272s & 391s \\
gpt-5.4           & 1384 & 1207 & 9s & 8s \\
Qwen2.5-72B       &  633 & 17565 & 9s & 168s \\
\bottomrule
\end{tabular}
\end{table}

When the responses on the largest grids are split by outcome, the failed responses are the longer and slower ones for four of the five models. Kimi-K2.5 spends 391 seconds on average on an instance it gets wrong against 272 on one it gets right, and claude-opus-4-6 3598 characters against 2714. Only gpt-5.4 spends marginally more on its successes. The additional output does not make success more likely. We leave to future work repeating the Plan track with extended reasoning enabled, to test whether a larger test-time budget moves the horizon $L_{50}$.

\subsection{Detailed Metrics}
\label{apx:metrics}

Fix an experimental cell $c=(\text{model},\text{level},\text{seed})$. At
temperature $t=0.7$, each cell has $S=3$ independent samples indexed by $i\in\{1,2,3\}$.

\subsubsection*{PREDICT}

Let $\hat{s}_c^{(i)}=\big((\hat x,\hat y),\hat d\big)$ be the LLM's predicted
final state and $s^\star_c=\big((x^\star,y^\star),d^\star\big)$ the true final
state (env-simulated).
\[
  \text{success}_c^{(i)} \;=\; \mathbb{1}\!\big[\hat{s}_c^{(i)} = s^\star_c\big]
  \qquad
  \text{accuracy}_c^{(i)} \;=\; \text{success}_c^{(i)}
\]
\[
  \text{manhattan\_distance}_c^{(i)} \;=\;
  d_M\!\big((\hat x,\hat y),(x^\star,y^\star)\big)
\]

\subsubsection*{PLAN}

Let:
\begin{itemize}
  \item $\mathcal{A}_c^{\text{bot}}$, $\mathcal{A}_c^{\text{llm},(i)}$ --- action sequences (bot's and LLM's, respectively)
  \item $\tau_c^{\text{bot}},\ \tau_c^{\text{llm},(i)}\in\{0,1\}$ --- \texttt{terminated} flag from env after executing each sequence
  \item $p_c^{\text{llm},(i)}=(x^{\text{llm}},y^{\text{llm}})$ --- LLM agent's final position
  \item $p_c^{\text{ball},(i)}$ --- red ball position (may be absent if the agent picked it up / it left the grid)
\end{itemize}
\[
  \text{SR}_c^{(i)} \;=\; \tau_c^{\text{llm},(i)}
\]
\[
  \text{number\_of\_bot\_actions}_c \;=\; \big|\mathcal{A}_c^{\text{bot}}\big|,
  \qquad
  \text{number\_of\_llm\_actions}_c^{(i)} \;=\; \big|\mathcal{A}_c^{\text{llm},(i)}\big|
\]

\[
  \text{llm\_efficiency}_c^{(i)} \;=\;
  \dfrac{\,\big|\mathcal{A}_c^{\text{bot}}\big|\,}{\,\big|\mathcal{A}_c^{\text{llm},(i)}\big|\,}
  \qquad \text{defined only when } \tau_c^{\text{llm},(i)}=1 \text{ and } \big|\mathcal{A}_c^{\text{llm},(i)}\big|>0
\]
Efficiency is undefined on failed samples, and all reported efficiency aggregates are means over successful samples only (efficiency conditional on success). Because the OmniBot reference is near-optimal rather than provably optimal, the ratio can occasionally exceed 1.

\[
  \text{llm\_ball\_distance}_c^{(i)} \;=\;
    \max\!\Big(0,\ d_M\!\big(p_c^{\text{llm},(i)},\,p_c^{\text{ball},(i)}\big)-1\Big)
\]

\subsubsection*{DECOMPOSE}

Let:
\begin{itemize}
  \item $\delta_c^{(i)}\in\{0,1\}$ --- whether the bot finished the mission while executing the LLM's subgoal plan
  \item $N_{\text{help},c}^{(i)}$ --- number of extra subgoals the bot injected at runtime beyond the LLM's plan
  \item $M_{\text{help},c}$ --- oracle's from-scratch decomposition size (independent of model/sample, depends only on level + seed)
\end{itemize}
\[
  \text{CR}_c^{(i)} \;=\; \delta_c^{(i)}
\]
\[
  \text{PR}_c^{(i)} \;=\; \delta_c^{(i)}\cdot\mathbb{1}\!\big[N_{\text{help},c}^{(i)}=0\big]
\]
\[
  \text{ACI}_c^{(i)} \;=\;
  \begin{cases}
    \max\!\Big(0,\ \dfrac{M_{\text{help},c}-N_{\text{help},c}^{(i)}}{M_{\text{help},c}}\Big)
      & \text{if } \delta_c^{(i)}=1 \text{ and } M_{\text{help},c}>0 \\[10pt]
    1 & \text{if } \delta_c^{(i)}=1 \text{ and } M_{\text{help},c}=0 \\[4pt]
    0 & \text{if } \delta_c^{(i)}=0
  \end{cases}
\]
\end{document}